# A welding penetration prediction model for laser welding process based on self-supervised learning using physics-informed neural networks

Sen Li[a], Xiaoying Liu[a], Xiaojian Xu[a], Chendong Shao[a], Yaqi Wang[a], Ling Lan[a,b], Xinhua Tang[a], Haichao Cui[a,*]

[a] *Shanghai Key Laboratory of Materials Laser Processing and Modification, School of Materials Science and Engineering, Shanghai Jiao Tong University, Shanghai 200240, PR China.*

[b] *Shanghai Shipbuilding Technology Research Institute, Shanghai 200032, China.*

**Abstract:**

The laser welding full-penetration is of critical importance, as it constitutes one of the fundamental factors in achieving defect-free welded joints. Accurate prediction of the penetration state is therefore essential for ensuring weld quality. To this end, this paper introduces SimPhysNet, a novel algorithm that achieves high classification accuracy in laser welding penetration prediction using only a limited number of labelled images. This approach effectively overcomes the limitations of supervised learning classification algorithms, which are hindered in industrial applications by their dependence on extensive, high-quality labelled data. The core of SimPhysNet is a unique self-supervised learning paradigm that embeds physical priors into a contrastive learning framework. By incorporating a physics-informed neural network (PINN), the model is guided to extract physically meaningful features of the molten pool and keyhole from a large set of unlabelled data, while three image augmentation tasks further enhance its generalization capabilities. Subsequently, a few-shot learning strategy, based on prototypical networks, enables robust classification by constructing class representations from a minimal set of labelled images. Experimental results demonstrate that SimPhysNet achieves a classification accuracy of 96.06% using only 200 labelled images (approximately 5% of the total labelled dataset), which is comparable to the performance of conventional supervised learning algorithms that utilize the entire labelled dataset. This work presents a new, efficient, and highly accurate method, providing the way for the intelligent automation of laser welding.

**Keywords:** Laser welding; Penetration prediction; Self-supervised learning; Physics-informed neural network

---

[*] Corresponding author: haichaocui@sjtu.edu.cn

## 1. Introduction

In recent years, laser welding technology has been extensively applied across various industries such as nuclear energy, shipbuilding, aerospace, defense equipment, automotive manufacturing, and railway vehicle production[1]. Compared to conventional welding techniques, laser welding offers significant advantages, including low thermal input, high weld depth-to-width ratios, rapid welding speed, and a high level of automation. Consequently, it is regarded as one of the most promising advanced joining technologies of the 21st century. However, achieving high-quality weld joints in laser welding requires precise control and optimization of process parameters. Therefore, the development of a reliable model for the entire laser deep penetration welding process would enable real-time adjustment of welding parameters to ensure full penetration and superior joint quality.

During the laser welding process, a substantial amount of signals are generated, primarily including acoustic and optical signals [2]. In recent years, optical signals have attracted growing attention from researchers due to their rich information content, ease of acquisition, and strong interpretability. For instance, Kang et al. [3] developed a deep learning model using optical emission spectroscopy (OES) to enable real-time prediction of the cross-sectional geometry of welds in aluminum/copper laser dissimilar welding (LDW) processes, thereby facilitating an intuitive assessment of welding quality. Similarly, Li et al. [4] designed an online monitoring system for laser weld geometry based on a cost-effective charge-coupled device (CCD) camera and a multi-task convolutional neural network, achieving real-time prediction of weld depth and width with high accuracy. However, despite their impressive accuracy, these deep learning models are predominantly supervised, creating a significant barrier to their widespread industrial application. The deployment of such data-driven models in industrial settings is frequently hindered by a significant practical challenge: the prohibitive cost associated with acquiring large-scale, high-quality labeled datasets. In the context of laser welding, establishing the ground truth for penetration state is fundamentally a labor-intensive, post-process activity. Each label necessitates careful inspection of the completed weld bead—typically by examining its backside—to confirm full penetration. When this verification process must be scaled to thousands or tens of thousands of samples to adequately cover the diverse range of process parameters and material conditions, it becomes a critical bottleneck. This reliance on extensive post-process verification impedes the rapid development and iteration of models and limits their scalability in dynamic production environments. Therefore, developing a methodology that can achieve high prediction accuracy with minimal reliance on labeled data is of paramount importance for the intelligent automation of laser welding.

To address these challenges, researchers have proposed Few-Shot learning methods. Few-Shot learning aims to achieve efficient learning with a limited number of samples, thereby reducing reliance on large labeled datasets. One of the most significant approach in this field is metric-based learning [5]. This approach involves training a model to map input samples into a low-dimensional embedding space, where samples from the same class are closely clustered, while those from different classes are distinctly separated. During inference, the class of a query sample is determined by calculating its distance to samples in the support set within this embedding space. Raffin et al. [6] found that Prototypical Networks exhibit outstanding performance under conditions of limited data availability, achieving an accuracy of 94.4% in quality monitoring tasks related to data scarcity in hook-type winding laser welding. This provides a data-efficient solution for industrial visual inspection. Addressing the challenge of limited small-sample data in industrial weld seam inspection, Zhu et al. [7] proposed a metric learning approach based on a distance metric classification mechanism, which enhances detection accuracy under extremely constrained data conditions. Liu et al. [8] introduced a novel algorithm called FS-Classifier, which employs unsupervised learning for pre-training the feature extractor and integrates it with an optimized prototypical network for small-sample classification. This method achieved high-precision classification of gas tungsten arc welding defects using small-sample datasets, further demonstrating the feasibility of combining unsupervised pre-training with prototypical network fine-tuning. While these works demonstrate the potential of combining unsupervised pre-training with few-shot learning, their pre-training objectives are purely data-driven. For complex physical processes like laser welding, such approaches may fail to capture the underlying causal relationships, instead learning superficial visual correlations. This limits their ability to generalize robustly from a feature space that is not grounded in the process physics.

In recent years, integrating physical information into deep learning networks has attracted significant attention. Physics-informed neural networks (PINNs) incorporate physical laws as prior knowledge into the neural network training process. This approach ensures that while the model learns from data, it remains consistent with fundamental physical principles. Consequently, PINNs can leverage physical constraints to guide the learning of accurate data features using limited experimental datasets, aligning with the goal of achieving high classification accuracy with fewer samples. Chen et al. [9] proposed a hybrid CNN approach that integrates physical prior knowledge by converting welding-related physical laws into image features that inform the deep learning process, thereby significantly improving the accuracy of variable polarity plasma arc (VPPA) welding quality prediction. Similarly, Lu et al.[10] developed a physics-informed CNN-LSTM (PI-CNN-LSTM) for real-time monitoring of laser filler-wire welding. Their architecture couples spatiotemporal molten pool features with an analytical heat transfer model to infer thermal-related variables. By employing physics-aware activation functions to constrain these parameters within meaningful ranges, they achieved robust penetration depth prediction even when trained on only 10% of the dataset. Li et al. [11] introduced an improved M-PINN for predicting the fatigue life of double-sided welds by incorporating implicit physical models within activation functions and explicit physical constraints into loss functions, which significantly enhance prediction accuracy under conditions of limited data availability. Although PINNs have shown success in supervised contexts for solving forward and inverse problems, their potential as a regularizer within self-supervised feature learning frameworks remains largely untapped. Existing studies typically use PINNs to approximate a specific solution. In contrast, our work re-purposes this paradigm to enforce physical consistency on a high-dimensional feature representation, a fundamentally different and novel application.

To address the aforementioned challenges, this paper introduces SimPhysNet, a novel framework designed to achieve high-accuracy penetration prediction from a limited set of labeled images. The primary contributions of this work are summarized as follows:

1. A novel fusion of physics and self-supervision for feature learning: Unlike conventional self-

supervised methods that rely solely on data augmentation, our framework pioneers the integration of physical priors, governed by partial differential equations, are embedded directly into a contrastive learning loss function. This approach is distinct from prior works, as it utilizes fundamental physical laws as an implicit supervisory signal to guide the feature extractor in learning physically meaningful representations from unlabeled data, thereby ensuring that the learned features are not merely correlational but are grounded in the underlying process dynamics.

2. Synergistic integration of pre-training and few-shot learning: The model architecture is structured in two stages. An initial pre-training stage leverages unlabeled data to establish a robust and generalized feature space. Subsequently, a fine-tuning stage employs a prototypical network-based few-shot learning strategy to achieve high classification accuracy with minimal labeled data. This synergistic design effectively mitigates the overfitting problem commonly encountered in data-limited scenarios.

3. Application-tailored image augmentation strategies: Three supervised pretext tasks—rotation prediction, Gaussian blur level prediction, and random cropping localization—are designed and integrated into the pre-training stage. These tasks compel the network to learn hierarchical features, from global molten pool morphology to fine-grained local details, which are critical for discriminating between penetration states and enhancing the model's overall generalization capabilities.

# 2. Materials and methods

## 2.1 Experimental setup

The experimental setup primarily comprises three key components: the laser welding module, the robotic motion control system, and the coaxial image acquisition module. The laser welding module consists of a fiber laser welding machine, a water-cooling unit, and a laser welding head. During the welding process, the workpiece remains stationary, while a Yaskawa robot arm precisely controls the movement of the welding head. Prior to initiating the welding procedure, the 808 nm auxiliary laser light source and the focal length of the coaxial camera are carefully adjusted to ensure the acquisition of a clear, stable, and well-illuminated image. Throughout the welding operation, images are captured and transmitted in real-time to a connected PC for data storage. The output power of the fiber laser is controlled by the PC, whereas other critical welding parameters, such as welding speed and defocus distance, are controlled via the robotic system.

The laser employed in this study is the YLS-20000 fiber laser with a spot-ring configuration, capable of delivering a maximum output power of 20 kW. The ring laser has a beam parameter product (BPP) of $15.377\ mm \cdot mrad$, with a focused spot diameter of 596.364 μm and a rayleigh length of 26.25 mm. The dot laser exhibits a BPP of $3.385\ mm \cdot mrad$, a focused spot diameter of 197.535 μm, and a rayleigh length of 2.88 mm. A schematic illustration of the laser welding setup is shown in **Fig. 1**. In the experimental configuration, a coaxial complementary metal oxide semiconductor (CMOS) camera was utilized to capture top-down images of the molten pool. An 808 nm auxiliary light source was integrated into the imaging system. During image acquisition, both the CMOS camera and the auxiliary light source were synchronously triggered by the control PC, ensuring stable illumination and improved image quality. The imaging system employs a $1/4''$ monochrome CMOS sensor positioned at a $0.0°$ angle. An 808 nm narrow bandpass filter (NBF) with a full width at half maximum (FWHM) of 30 nm is integrated into the setup. Key camera parameters were set with an exposure time of $36$ μs, a gamma value of 1.2, and a black level of 1.0%.

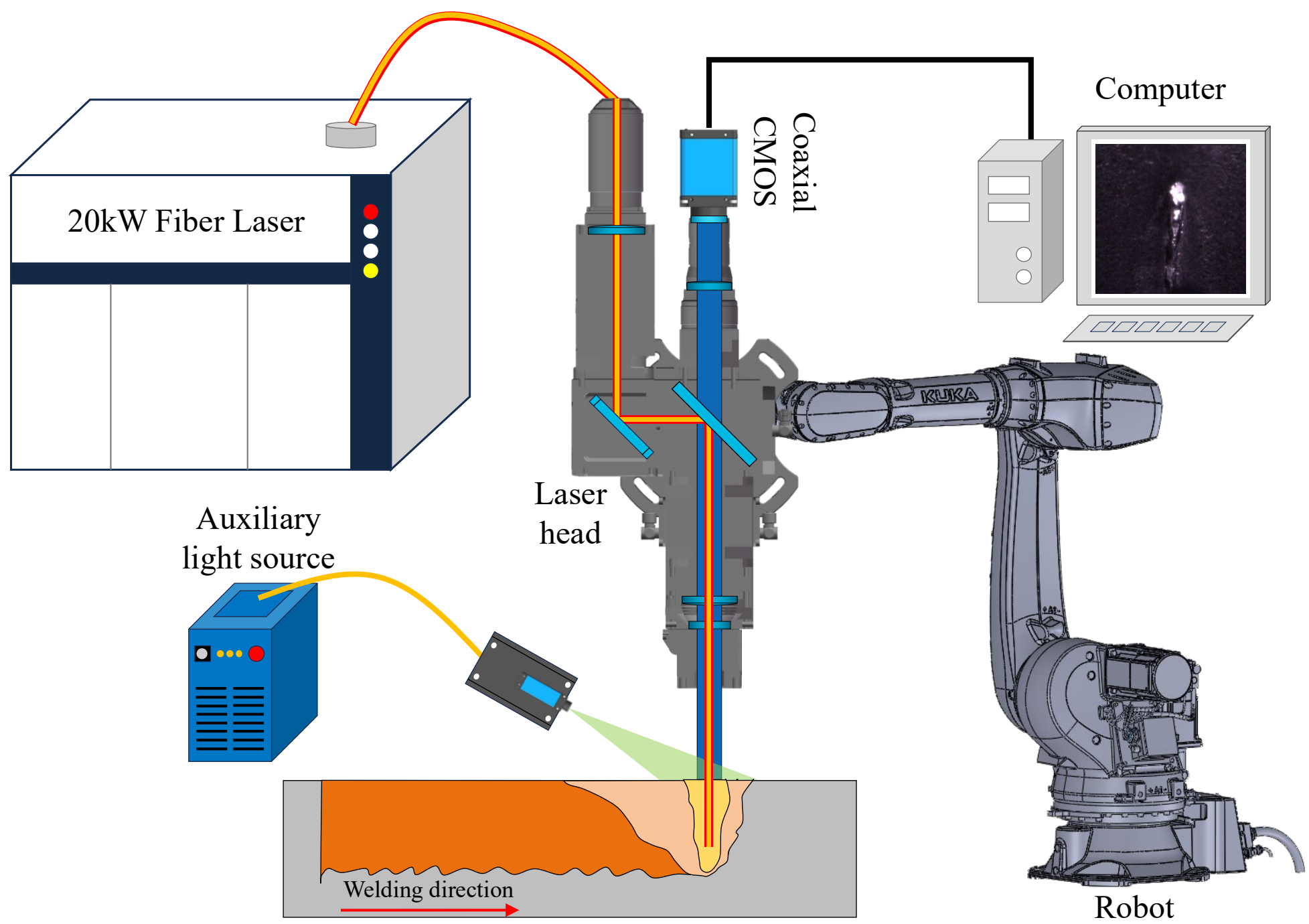

**Fig. 1** The schematic of the welding system.

## 2.2 Experimental method

The welding material utilized is COST-E, with dimensions of $100\,mm \times 80\,mm$ (length × width) as illustrated in **Fig. 2**. The specimens are provided in five different thicknesses: 3.0 mm, 4.0 mm, 6.0 mm, 8.0 mm, and 10.0 mm, with chemical compositions detailed in **Tab. 1**. Prior to welding, the specimens are thoroughly cleaned using anhydrous alcohol to remove any contaminants that could potentially affect the weld joint quality. During the welding process, argon gas is used as a shielding gas to protect the weld area. The argon is supplied through three parallel copper tubes positioned behind the weld seam, with a flow rate of $25\,L/min$ and a purity of 99.99%. The welding parameters are summarized in **Tab. 2**. By varying the plate thickness $T[mm]$, welding speed $V_w[m/min]$, spot laser power $P_d[kW]$, and defocus amount $D_f[mm]$, various weld joint conditions are obtained.

**Tab. 1** Chemical composition of COST-E(wt.%) [12].

| COST-E | C | Mn | Si | Ni | Cr | Mo | W | Nb | V | Fe |
|---|---|---|---|---|---|---|---|---|---|---|
| (wt.%) | 0.12 | 0.45 | 0.1 | 0.75 | 10.5 | 1.0 | 1.0 | 0.05 | 0.2 | Bal. |

**Tab. 2** Welding parameters.

| Parameter | $P_d$ (kW) | $D_f$ (mm) | $V_w$ (m/min) | $T$ (mm) |
|---|---|---|---|---|
| Value | [3.0, 4.0, 5.0, 6.0, 7.0, 8.0, 9.0] | [0.0, -5.0] | [1.0, 1.5, 2.0] | [3.0, 4.0, 6.0, 8.0, 10.0] |

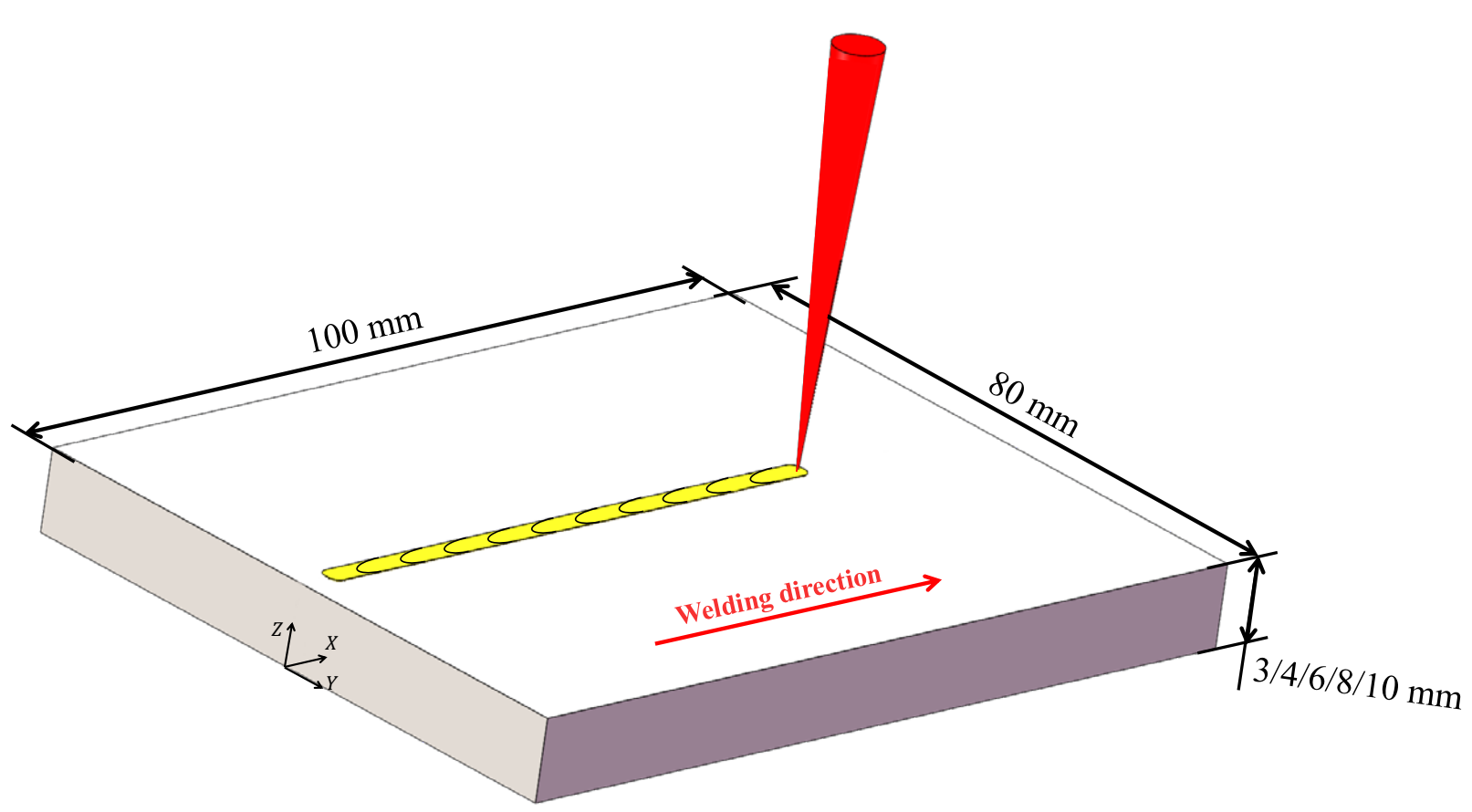

**Fig. 2** Schematics of the laser welding experiment.

### 2.3 Dataset acquisition

Throughout the welding process, a considerable number of molten pool images, representing both fully penetrated and non-penetrated welding states, were captured in real-time using the experimental setup described in Section 2.1. **Fig. 3** illustrates the front and rear surfaces as well as the cross-section of the weld after welding. The labeled dataset is denoted as $D_L = \{x_{i,L}, y_{i,L}\}_{i=1}^{N_L}$, and the unlabeled dataset is denoted as $D_U = \{x_{i,U}\}_{i=1}^{N_U}$, where $N$ denotes the total number of images in each respective dataset. Here, $x_{i,L}$ and $x_{i,U}$ denote the molten pool images, and $x_{i,L} \in \{1, \ldots, K\}$ represent the corresponding annotation labels, with $K$ indicating the total number of categories. In the context of the laser welding penetration dataset, $K = 2$. The unlabeled images were collected from routine welding experiments. The composition of the dataset is illustrated in **Tab. 3**, and representative samples are depicted in **Fig. 4**. Additionally, the following formula is used to calculate the model's accuracy:

$$Acc = \frac{TP + TN}{TP + TN + FP + FN} \times 100\% \quad (1)$$

where TP (true positive), FP (false positive), FN (false negative), and TN (true negative) represent the number of correctly and incorrectly classified samples.

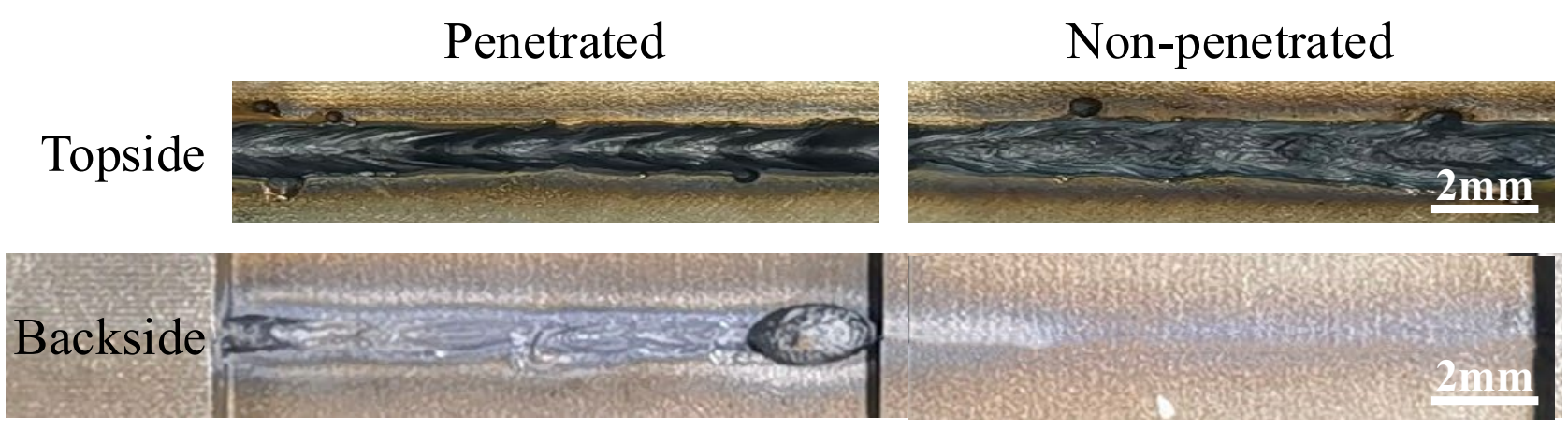

**Fig. 3** Schematic diagrams illustrating the topside and backside views of the weld.

**Tab. 3** Composition of the datasets. ("-" denotes the absence of the corresponding content)

| Datasets | Classes | Train set | Validation set |
|---|---|---|---|
| $D_L$ | Penetrated (Class 0) | 1400 | 600 |
| | Non-penetrated (Class 1) | 1400 | 600 |
| $D_U$ | - | 8000 | - |

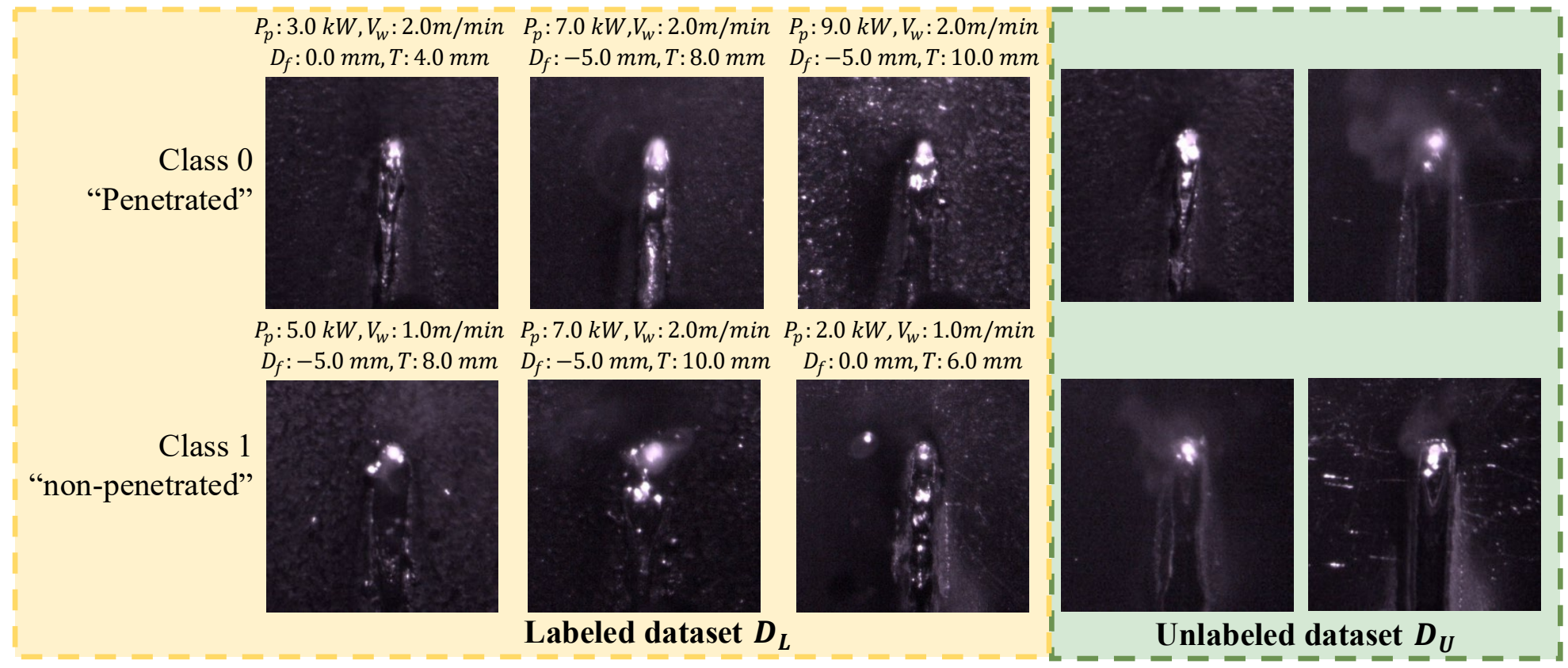

**Fig. 4** The samples of labeled dataset $D_L$ and unlabeled dataset $D_U$.

## 3. Details of the network

As discussed in the first section, a laser molten pool classification model with strong generalization performance and high accuracy typically depends on large-scale, high-quality datasets. However, the annotation of such datasets is both labor-intensive and time-consuming. To address this issue, this paper proposes an unsupervised classification network, termed SimPhysNet, which achieves substantial classification accuracy using only a limited number of labeled samples. **Fig. 5** presents the detailed architecture of the proposed network, which consists of two key components: fist, a self-supervised learning strategy is applied to train the backbone network, enabling it to efficiently extract informative features from molten pool images. Second, a few-shot classification method is utilized to train the feature classifier. Notably, to enhance the effectiveness of the self-supervised learning phase, a PINN is incorporated into the training process.

The fundamental hypothesis underpinning the SimPhysNet architecture is that for complex physical systems such as laser welding, optimal feature representations must be consistent with governing physical principles. A purely data-driven model risks learning spurious correlations from image data, particularly when labeled samples are scarce. Consequently, the core methodological innovation of SimPhysNet lies in the fusion of a data-driven contrastive learning objective with physics-based constraints imposed by a PINN module. The PINN functions as a form of regularization, ensuring that the feature embedding space aligns with the laws of heat conduction and energy distribution, thereby enhancing the model's robustness and generalization performance without requiring additional annotated data.

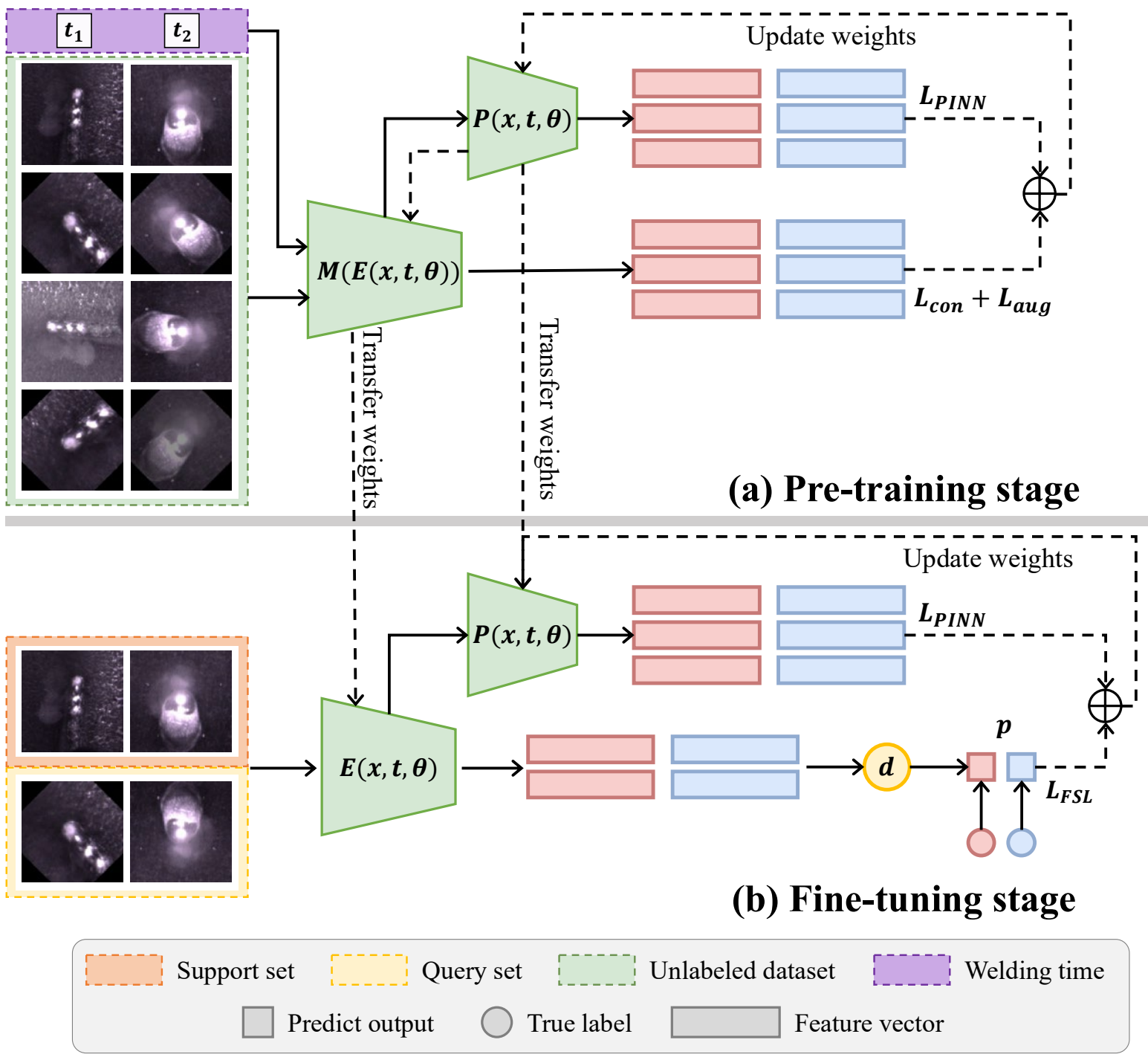

**Fig. 5** The overall architecture of the proposed network in a 2-way 1-shot paradigm. The model embedding backbone is denoted by $M(E(x,t,\theta))$ and euclidean distance function is denoted by $d$. $L_{con}$, $L_{aug}$ and $L_{PINN}$ denote the contrastive loss, image task loss and PINN loss in the pre-training stage, $L_{FSL}$ denote the few-shot loss in the fine-tuning stage. $p$ denotes the predicted logits/output from the model.

## 3.1 Pre-training stage

### 3.1.1 Contrastive learning

Self-supervised learning methods typically operate on large volumes of unlabeled images. These methods do not rely on manually annotated category labels; instead, they leverage the data itself as supervisory signals to learn feature representations, which can subsequently be utilized in downstream tasks. Specifically, the contrastive learning approach adopted in this study aims to train an encoder that generates similar representations for instances within the same class, while maximizing the dissimilarity between the representations of different classes. By adhering to this principle, the model is capable of extracting meaningful features from extensive unlabeled datasets, thereby significantly enhancing its generalization performance and robustness.

**Fig. 6** illustrates the core implementation framework of the Simsiam algorithm[13]. Prior to being fed into the network, the molten pool image $x$ undergoes random data augmentation, resulting in two transformed views, $x_1$ and $x_2$. These augmented images are then processed through the encoder network $E(\cdot)$. The encoder network comprises a backbone architecture and a projection head implemented as a MLP, where ResNet-18[14] is employed as the backbone network in this work. Following the $E_1(\cdot)$ encoder, an additional MLP-based prediction head $M(\cdot)$ is introduced. This prediction network maps the high-dimensional feature representation generated by $E_1(\cdot)$ to align with the feature space produced by $E_2(\cdot)$. Consequently, the transformed outputs for the two augmented images are expressed as $r_1 = M(E_1(x_1))$ and $k_2 = E_2(x_2)$. Evidently, the optimization objective of the network is to maximize the similarity between $q_1$ and $k_2$. To achieve this, the model minimizes their negative cosine similarity:

$$L_D(k_1, q_2) = -\frac{k_1}{\|k_1\|^2} \cdot \frac{q_2}{\|q_2\|^2} \tag{2}$$

Where $\|\cdot\|^2$ denotes the L2 norm. Furthermore, the loss function for contrastive learning can be defined as:

$$L_{con} = \frac{1}{2} L_D(k_1, q_2) + \frac{1}{2} L_D(k_2, q_1) = -\left(\frac{k_1}{2\|k_1\|^2} \cdot \frac{q_2}{\|q_2\|^2} + \frac{k_2}{2\|k_2\|^2} \cdot \frac{q_1}{\|q_1\|^2}\right) \tag{3}$$

An important optimization of this contrastive learning approach is the application of the stop-gradient (stopgrad) operation. This enables the network to achieve effective learning outcomes without relying on negative sample pairs or momentum encoders. Accordingly, Equations (1) and (2) are reformulated as follows:

$$L_D(k_1, stopgrad(q_2)) \tag{4}$$

$$L_{con} = \frac{1}{2} L_D(k_1, stopgrad(q_2)) + \frac{1}{2} L_D(k_2, stopgrad(q_1)) \tag{5}$$

Here, $stopgrad(\cdot)$ signifies the stop-gradient operation. Equation (4) defines the loss function of the Simsiam algorithm.

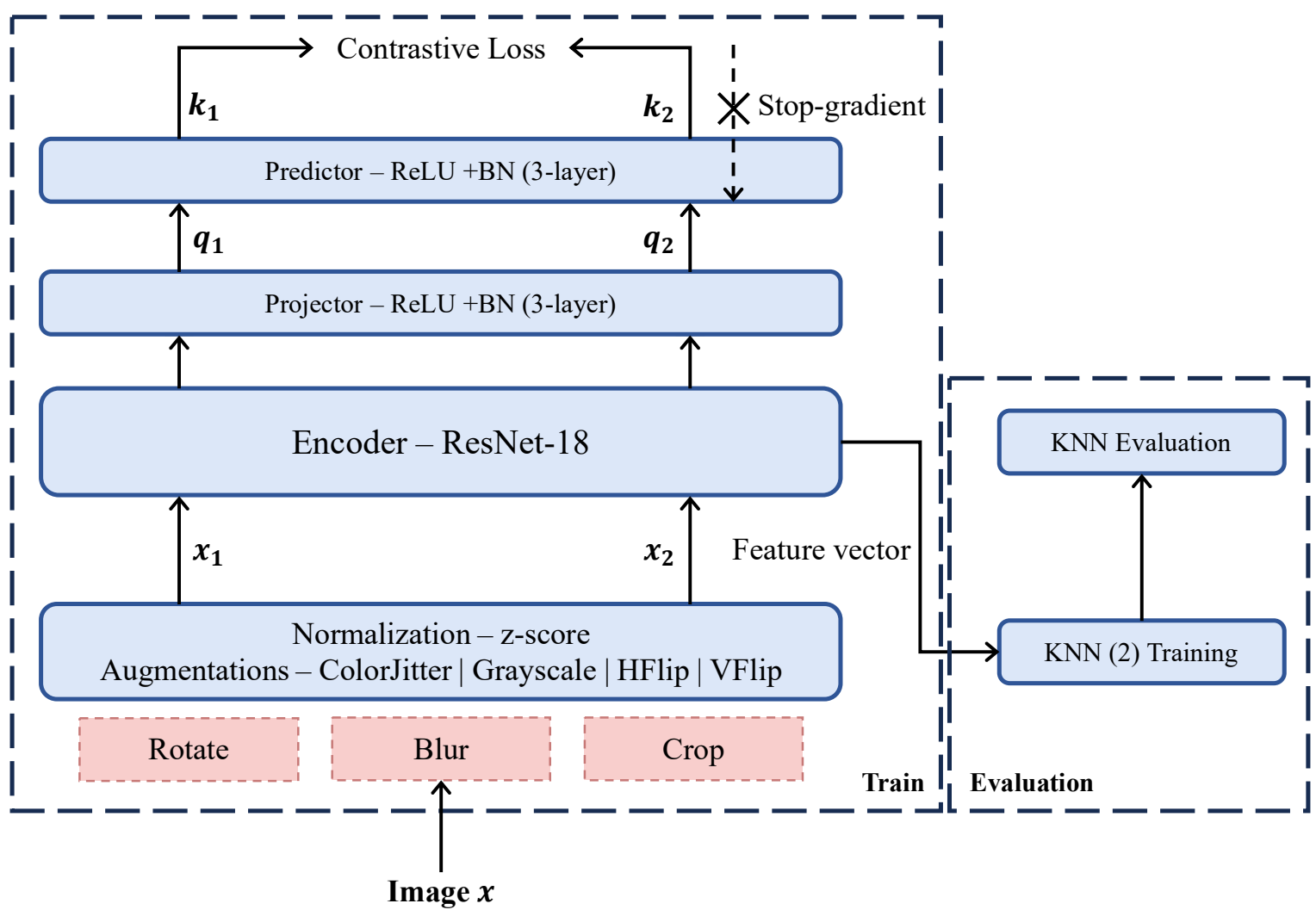

**Fig. 6** Contrastive Learning process design.

To effectively capture the high-dimensional features of the laser welding melt pool, this study introduces three branched prediction heads designed to perform distinct image augmentation tasks. The first task involves predicting the rotation angle, where the input image is randomly rotated by a predefined angle $\alpha$. The second task consists of applying random Gaussian blur. The third task involves random cropping, wherein the input image is cropped to a resolution of $224 \times 244$ pixels. The prediction head responsible for estimating the rotation angle is denoted as $A_1(\cdot)$, the prediction head for Gaussian blur parameters is denoted as $A_2(\cdot)$, and the prediction head for random cropping is denoted as $A_3(\cdot)$. Consequently, the aforementioned image augmentation tasks are achieved by optimizing the following loss function.

$$\begin{aligned} L_{aug} &= \frac{1}{2}L_{aug}(x_1) + \frac{1}{2}L_{aug}(x_2) \\ &= \frac{1}{2}\Big(\alpha_1 L_{Rot}(x_1, \theta) + \beta_1 L_{blur}\big(x_1, \sigma_c, \sigma_d, \sigma_x, \sigma_y\big) + \gamma_1 L_{crop}\big(x_1, C_x, C_y\big)\Big) \\ &+ \frac{1}{2}\Big(\alpha_2 L_{Rot}(x_2, \theta) + \beta_2 L_{blur}\big(x_2, \sigma_c, \sigma_d, \sigma_x, \sigma_y\big) + \gamma_2 L_{crop}\big(x_2, C_x, C_y\big)\Big) \end{aligned} \tag{6}$$

$$\alpha_1 + \beta_1 + \gamma_1 = 1, \alpha_2 + \beta_2 + \gamma_2 = 1 \tag{7}$$

Here, $L_{Rot}$, $L_{blur}$, and $L_{crop}$ denote the loss functions corresponding to the angle prediction task, Gaussian blur prediction task, and random cropping task, respectively. The coefficients $\alpha$, $\beta$, and $\gamma$ are used to balance the contribution of these three loss terms. $\theta$ represents the random rotation angle applied to the image, while $\sigma_x$ and $\sigma_y$ indicate the coordinates of the point with the highest Gaussian blur intensity in the image, $\sigma_c$ denotes the magnitude of the blur at that point, and $\sigma_d$ is the decay rate of the Gaussian blur intensity. $C_x$ and $C_y$ are the coordinates of the center of the cropped image within the original image, with the cropped region having a fixed size of $224 \times 224$ pixels.

Furthermore, the root mean square error (RMSE) function is employed as the loss function for the aforementioned processes. Therefore, the loss function for the angle prediction task can be written as follows:

$$L_{Rot}(x_1, \theta) = \left(\frac{1}{n}\sum_{i=1}^{n} \|\theta_i - \hat{\theta}_i\|^2\right)^{\frac{1}{2}} \tag{8}$$

Here, $\theta_i$ represents the actual rotation angle, and $\hat{\theta}_i$ denotes the predicted value of $A_1(\cdot)$.

Similarly, the loss functions for the Gaussian blur prediction task and the random cropping task can be expressed as follows:

$$\begin{aligned} L_{blur}\big(x_1, \sigma_c, \sigma_d, \sigma_x, \sigma_y\big) &= \left(\frac{1}{n}\sum_{i=1}^{n} \|\sigma_c - \hat{\sigma}_c\|^2\right)^{\frac{1}{2}} + \left(\frac{1}{n}\sum_{i=1}^{n} \|\sigma_d - \hat{\sigma}_d\|^2\right)^{\frac{1}{2}} \\ &+ \left(\frac{1}{n}\sum_{i=1}^{n} \|\sigma_x - \hat{\sigma}_x\|^2\right)^{\frac{1}{2}} + \left(\frac{1}{n}\sum_{i=1}^{n} \|\sigma_y - \hat{\sigma}_y\|^2\right)^{\frac{1}{2}} \end{aligned} \tag{9}$$

$$L_{crop}\big(x_2, C_x, C_y\big) = \left(\frac{1}{n}\sum_{i=1}^{n} \|C_x - \hat{C}_x\|^2\right)^{\frac{1}{2}} + \left(\frac{1}{n}\sum_{i=1}^{n} \left\|C_y - \hat{C}_y\right\|^2\right)^{\frac{1}{2}} \tag{10}$$

Both take forms similar to Equation (8) and will not be elaborated further here.

In the task of classifying the top surface molten pool in laser welding, the morphological characteristics of the molten pool are the primary focus of the network. Additionally, we recognize that this aspect contributes to image enhancement. Based on these two premises, we have formulated and designed three tasks.

Firstly, the image rotation operation compels the model to learn orientation-invariant global features rather than relying on position-fixed local features. This suggests that the network must analyze the global structures within the image to determine the applied rotational transformation.

Given the context of laser welding penetration classification, where image content is relatively simple and the primary targets are the molten pool and small pores, the network can focus on the morphology of these elements to predict the rotation angle.

Secondly, Gaussian blurring reduces image details such as textures and edges, thereby forcing the model to learn more generalized global features and avoid over-fitting to specific sharp features present in the training set. Moreover, during actual image acquisition, the prevalence of blurring is a direct consequence of real-world industrial laser welding conditions. Specifically, blurs originate from physical phenomena such as metal vapor and fume interference, plasma formation, rapid molten pool dynamics that can induce motion blur, and minor focus drift or mechanical vibrations from the robotic arm during prolonged operation. These factors introduce optical obstructions or motion artifacts. By simulating such conditions through data augmentation, the model's robustness and adaptability to real production environments are significantly improved.

Thirdly, by randomly cropping local regions of the image, the model is compelled to extract discriminative features from limited pixel areas, thereby increasing its sensitivity to fine-grained details. During molten pool classification, this encourages the model to focus on key indicators such as the presence or absence of keyholes, the morphology of molten pool edges, and subtle features such as spatter, enabling it to distinguish minor differences between similar regions.

In summary, the design of these three aforementioned tasks ensures that the feature vectors ultimately captured by the model contain both global and detailed features. Moreover, this approach can be regarded as a form of image enhancement, contributing to improved model generalization and robustness.

3.1.2 Physics informed neural network

The problem that PINN aim to address can be formulated as:

$$N[u(t,x,y,z)|\lambda] = 0, x \in \mathbb{R}^D, t \in \mathbb{R} \tag{11}$$

where, $N[u(t,x,y,z)|\lambda]$ denotes the differential operator used to describe the real physical system, and $u(t,x,y,z)$ represents the solution of the system. The primary function of PINN is to estimate the system's solution $u(t,x,y,z)$ using a deep neural network $D(t,x,y,z|\theta)$. It is noteworthy that this network provides only an approximate solution to $u(t,x,y,z)$. Let $\boldsymbol{X} = (x,y,z)$ represent the spatial coordinates. During training, the governing equation (11) acts as a physical constraint. The optimization objective is to minimize a loss function composed of the data misfit and the PDE residual. Let $\{t_i, \boldsymbol{X}_i\}_{i=1}^{N_d}$ be the set of training data points on the initial or boundary conditions, and $\{t_j, \boldsymbol{X}_j\}_{j=1}^{N_p}$ be the set of collocation points sampled within the physical domain. The composite loss function can be expressed as:

$$L_{PINN} = \left(\frac{1}{N_d}\sum_{i=1}^{N_d}\|u(t_i,\boldsymbol{X}_i) - D(t_i,\boldsymbol{X}_i)\|^2\right)^{\frac{1}{2}} + \left(\frac{1}{N_p}\sum_{j=1}^{N_p}\left\|N\left[D\left(t_j,\boldsymbol{X}_j\right)|\lambda\right]\right\|^2\right)^{\frac{1}{2}} \tag{12}$$

Here, the first term enforces the model to match the known data, while the second term enforces it to satisfy the governing physical law. It is evident that Equation (12) can also be regarded as a variant of the RMSE By explicitly embedding the physical governing equations into the construction of loss function, the PINN enables a progressive approximation of the deep learning model toward the true physical behavior. Compared to conventional numerical computation methods such as the finite element method (FEM) and the finite difference method (FDM), PINN does not rely on extensive and meticulously designed meshes. Instead, it employs automatic differentiation to update the model parameters, allowing the derivatives to be computed automatically during the training process. Essentially, this approach constructs a differentiable framework within the function space, leveraging the approximation capabilities of deep neural networks to transform the analytical solutions of physical equations into parameter optimization problems.

In SimPhysNet, our application of PINN leverages a specific network architecture to guide image feature learning by enforcing physical consistency. The process begins with the high-dimensional features extracted from molten pool images by the backbone encoder. These visual features are then fed into dedicated physical parameter prediction modules. These modules are designed to infer crucial physical parameters (such as the spatial boundaries of the welding zone and heat source characteristics) directly from the image features. Subsequently, a MLP approximates the temperature field $u(t,x,y,z)$ based on these image-inferred physical parameters. The PINN loss function then evaluates how well this feature-derived physical representation (the approximated temperature field) adheres to fundamental physical laws, specifically the heat conduction PDE and boundary conditions. By backpropagating this physical consistency loss through the network, the backbone encoder is implicitly forced to learn features that correspond to physically plausible molten pool states. This mechanism serves as a powerful regularization, guiding the model to extract features that are not merely visually correlated but are inherently consistent with the underlying physics, thereby enhancing generalization and robustness.

In this study, we utilize the partial differential equation (PDE) of heat conduction in a semi-infinite medium to develop a PINN capable of predicting the state of any sampling point at any given time during the welding process:

$$\frac{\partial T}{\partial t} - \alpha \nabla^2 T = 0 \tag{13}$$

where, $T$ is temperature and $t$ is time. $\alpha$ denotes the thermal diffusivity of the material, which is assumed to be a constant with a value of $1.17 \times 10^{-5}\ m^2 \cdot s^{-1}$ in our experiments. Therefore, the loss function for the thermal conduction term should be designed as follows:

$$L_{PDE} = \left( \frac{1}{N_p} \sum_{j=1}^{n} \left\| \frac{\partial D_{PDE}(t_j, \boldsymbol{X}_j)}{\partial t} - \alpha \nabla^2 D_{PDE}(t_j, \boldsymbol{X}_j) \right\|^2 \right)^{\frac{1}{2}} \tag{14}$$

The following equations are satisfied at the boundaries:

$$T(t = 0, x, y, z) = 26℃ \tag{15}$$

The loss function for the boundary condition term should be formulated as follows:

$$L_{IBC} = \left( \frac{1}{N_p} \sum_{j=1}^{n} \left\| T(t_j, \boldsymbol{X}_j) - D_{IBC}(t_j, \boldsymbol{X}_j) \right\|^2 \right)^{\frac{1}{2}} \tag{16}$$

In the context of welding, the heat conduction equation for the transient temperature field, known as the Fourier-Kirchhoff partial differential equation[15], is mathematically described as follows:

$$\begin{aligned} \rho c_p \cdot \frac{\partial T}{\partial t} &= \nabla \cdot (\lambda \nabla T) + q_v \\ &= \left[ \frac{\partial}{\partial x}\left(\lambda \frac{\partial T}{\partial x}\right) + \frac{\partial}{\partial y}\left(\lambda \frac{\partial T}{\partial y}\right) + \frac{\partial}{\partial z}\left(\lambda \frac{\partial T}{\partial z}\right) \right] + q_v \end{aligned} \tag{17}$$

where, $\rho$ denotes density, $c_p$ signifies specific heat, and $\lambda$ indicates thermal conductivity. The volumetric heat source density $q_v$ quantifies the amount of heat generated per unit time per unit volume of material, and it is utilized for modeling the heat source in the welding process.

In this study, the conical volumetric heat source model [15] with a Gaussian distribution is applied to simulate the heat source distribution during the laser welding process, as shown in **Fig. 7(a)**. The heat source energy density, $q_v|_{Con}\,[W \cdot m^{-3}]$, can be mathematically represented as follows:

$$q_v(x,y,z,t)|_{Con} = \frac{9\eta P e^3}{\pi(e^3-1)} \frac{1}{(z_e - z_i)(r_e^2 + r_e r_i + r_i^2)} exp\left( -\frac{3[(x - V_w t)^2 + y^2]}{r_0^2(z)} \right) \tag{18}$$

$$r_0(z) = r_e + \frac{r_i - r_e}{z_i - z_e}(z - z_e) \tag{19}$$

where, $P[W]$ is the laser power, $\eta[-]$ is laser absorption rate and $V_w[m \cdot s^{-1}]$ the welding speed. $r_e[m]$ and $r_i[m]$ represent the radii of the upper surface $z = z_e$ and lower surface $z = z_i$ of the heat source, respectively. $r_0$ is the heat source radius at the plane $z$, functioning as a linearly decreasing distribution from the upper surface to the lower surface. Based on the aforementioned analysis, the loss term associated with energy conservation of the conical heat source should be formulated as follows:

$$L_{Con} = \left( \frac{1}{N_p} \sum_{j=1}^{n} \left\| \rho c_p \frac{\partial D_{Con}(t_j, \boldsymbol{X}_j)}{\partial t} - \left( \nabla \cdot \left( \lambda \nabla D_{Con}(t_j, \boldsymbol{X}_j) \right) + q_v(t_j, \boldsymbol{X}_j) \right) \right\|^2 \right)^{\frac{1}{2}} \tag{20}$$

Furthermore, considering that a single heat source model may not fully characterize the thermal distribution in laser welding[16], this study employs a hybrid approach based on the Goldak's double ellipsoidal heat source model[17], as illustrated in **Fig. 7(b)**. The corresponding mathematical formulation is presented as follows:

$$q_{v1}(x,y,z,t)|_{Gol} = \frac{6\sqrt{3}f_1\Phi}{abc_1\pi^{3/2}} exp\left(\frac{-3(x-V_wt)^2}{a^2}\right) exp\left(-\frac{3y^2}{b^2}\right) exp\left(-\frac{3z^2}{c_1^2}\right) \tag{21}$$

$$q_{v2}(x,y,z,t)|_{Gol} = \frac{6\sqrt{3}f_1\Phi}{abc_2\pi^{3/2}} exp\left(\frac{-3(x-V_wt)^2}{a^2}\right) exp\left(-\frac{3y^2}{b^2}\right) exp\left(-\frac{3z^2}{c_2^2}\right) \tag{22}$$

$$f_1 = \frac{2c_1}{c_1+c_2}, f_2 = \frac{2c_2}{c_1+c_2}, f_1+f_2 = 2, q_v(x,y,z,t)|_{Gol} = q_{v1} + q_{v2} \tag{23}$$

where $\Phi$ is the effective laser power. $f_1$ and $f_2$ specify the proportion of energy distributed to the front and rear sections of the double ellipsoidal heat source. The semi-axis lengths $a$ and $b$ represent the width and depth of the heat source, respectively, while the semi-axis lengths $c_1$ and $c_2$ denote the lengths of the front and rear sections of the heat source, respectively. The loss term for Goldak heat source energy conservation model is consistent with Equation (20).

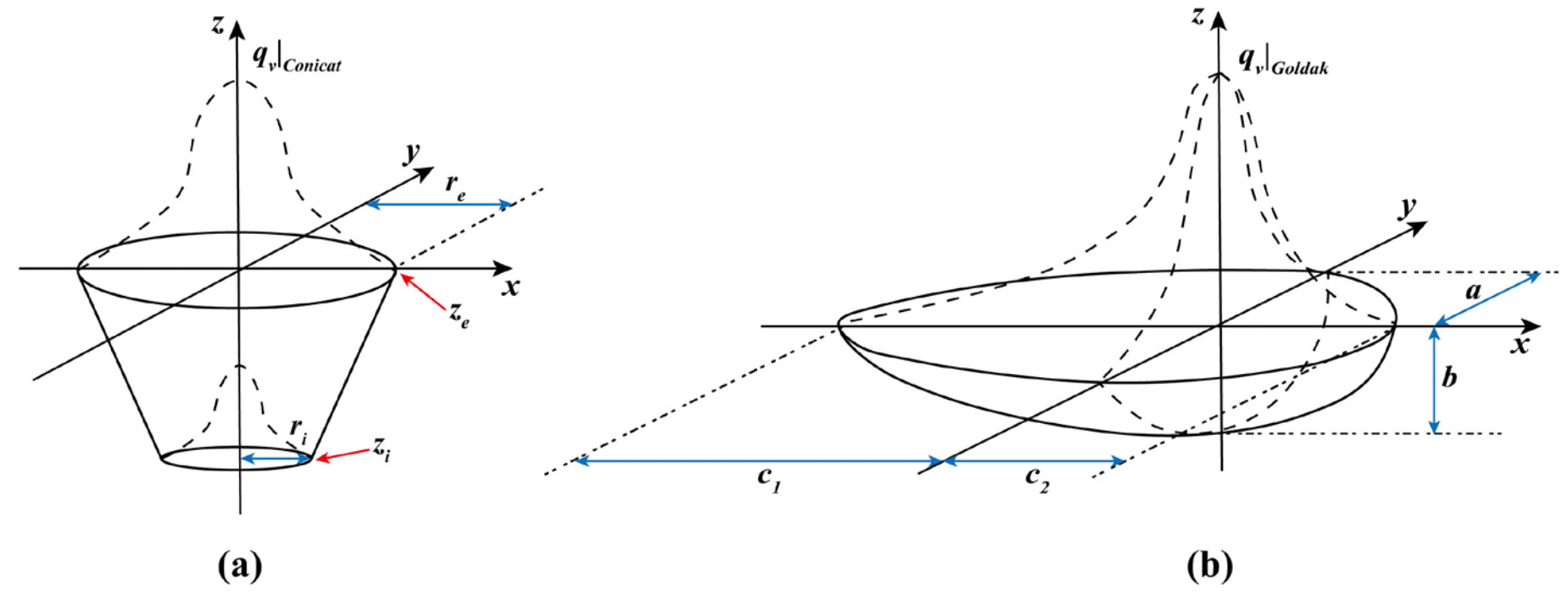

**Fig. 7** The forms of two types of welding heat sources: (a) Conical heat source model[15], (b)Goldak heat source model[17].

To prevent each term from being driven toward zero during the optimization process of the PINN, we have extended the RMSE algorithm by incorporating a penalty term, as exemplified in Equation (14):

$$L_{PDE} = \left(\frac{1}{n}\sum_{j=1}^{n}\left\|\frac{\partial D_{PDE}(t_j,\boldsymbol{X}_j)}{\partial t} - \alpha\nabla^2 D_{PDE}(t_j,\boldsymbol{X}_j)\right\|^2\right)^{\frac{1}{2}}$$
$$+\omega\cdot\left(\left(\left|\frac{\partial D_{PDE}(t_j,\boldsymbol{X}_j)}{\partial t}\right|+\varepsilon\right)^{-1} + \left(\left|\alpha\nabla^2 D_{PDE}(t_j,\boldsymbol{X}_j)\right|+\varepsilon\right)^{-1}\right) \tag{24}$$

In this manner, as each component approaches zero, the penalty term increases significantly, thereby steering the optimization direction away from zero. Here, $\omega$ denotes the penalty coefficient, which is consistently set to 0.05 in our experimental configuration. Additionally, $\varepsilon = 1e^{-6}$ is introduced as a regularization term to avoid division by zero.

A critical challenge in training PINNs is balancing the contributions of disparate loss terms ($L_{PDE}$, $L_{IBC}$, and $\lambda_{energy}$). Static weights require extensive manual tuning and may fail to adapt to the changing dynamics of the training process. To overcome this limitation, a hybrid adaptive weighting methodology is implemented. This is a sophisticated, multi-stage approach designed to autonomously regulate the influence of each loss component through three key stages: loss signal stabilization, normalization, and uncertainty-based dynamic weighting.

First, to mitigate the effects of high-frequency noise inherent in the stochastic optimization process, each loss signal is stabilized using an exponential moving average (EMA). The smoothed loss, $L_k^{smooth}$ , for each component $k \in \{PDE, IBC, energy\}$ at training step $t$ is calculated as:

$$L_k^{smooth(t)} = \alpha \cdot L_k^{smooth(t-1)} + (1-\alpha) \cdot L_k^t \tag{25}$$

where $L_k^t$ is the raw loss at the current step and $\alpha$ is a smoothing factor.

Second, to ensure all loss terms operate on a comparable numerical scale, the smoothed losses are normalized. Each smoothed loss is divided by its value at the beginning of training, $L_{k,initial}$, yielding a normalized loss $\hat{L}_k$:

$$\hat{L}_k = \frac{L_k^{smooth(t)} + \varepsilon}{L_{k,initial} + \varepsilon} \tag{26}$$

This normalization prevents any single term from dominating the total loss merely due to its magnitude.

Finally, the core of this method is the dynamic weight calculation, which combines a pre-defined initial weight, $w_{k,initial}$, with a learnable, uncertainty-based parameter, $s_k$. This approach allows for the incorporation of prior knowledge while enabling fine-grained, dynamic adjustments. The final weight $w_k$ for each loss is:

$$w_k = w_{k,initial} \cdot exp(-s_k) \tag{27}$$

The total loss function for the PINN is then formulated as the sum of the weighted, normalized losses and a regularization term on the learnable parameters. A ReLU function is applied to ensure non-negative loss contributions.

$$L_{PINN} = \sum_{k \in \{PDE, IBC, energy\}} \left( ReLU\left(\frac{1}{2} w_k \cdot \hat{L}_k\right) + |s_k| \right) \tag{28}$$

In this formulation, the model learns to down-weigh tasks with higher uncertainty (larger $s_k$), while the $|s_k|$ term regularizes the learning process to prevent any weight from collapsing to zero. This adaptive mechanism provides a robust and autonomous framework for balancing the multifaceted objectives in PINN training.

The architecture of the PINN is illustrated in **Fig. 8**. A multilayer perceptron (MLP)[18] is employed to approximate the solution $u(t, \boldsymbol{X})$. This approximated solution is then utilized to construct the heat conduction loss $L_{PDE}$, boundary condition loss $L_{IBC}$, and heat source loss $L_{energy}$. The parameters of the MLP are trained using a gradient descent optimization method based on the backpropagation of the loss function.

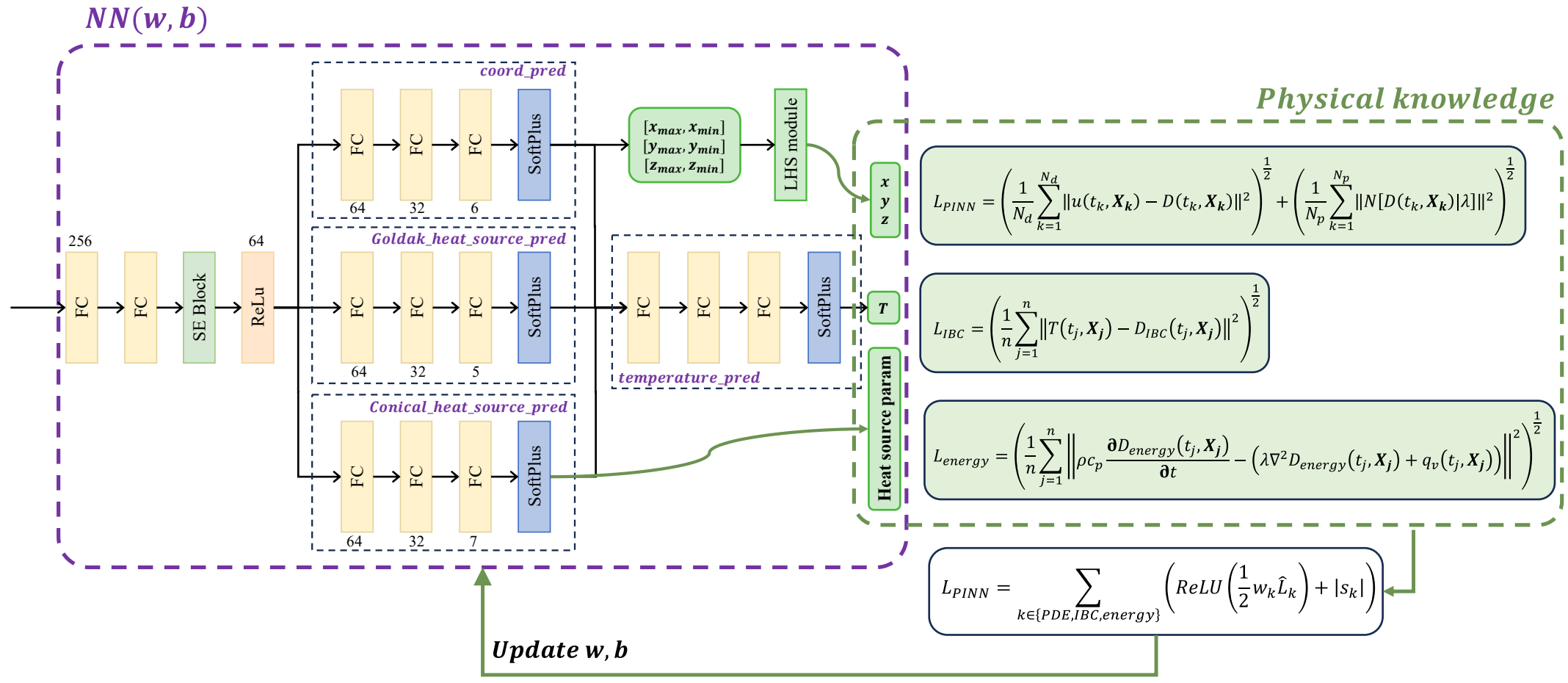

**Fig. 8** Overview of the PINN. A multilayer perceptron (MLP) is employed to approximate the solution $u(t, \boldsymbol{X})$, which is subsequently utilized to construct the heat conduction loss $L_{PDE}$ boundary condition loss $L_{IBC}$, and heat source loss $L_{energy}$. The derivatives of $u$ are computed using automatic differentiation in PyTorch. The parameters of the MLP are trained using a gradient descent optimization method based on the backpropagation of the loss function.

Latin hypercube sampling (LHS) is a highly efficient data sampling method[19]. Compared to conventional Monte Carlo sampling methods, LHS demonstrates superior performance in terms of sampling efficiency, reduced sampling variance, and lower correlation among multidimensional variables. By partitioning the parameter space into equally probable sub-intervals and selecting one sample point independently from each sub-interval, LHS effectively mitigates the clustering effect commonly observed in conventional random sampling approaches[20]. Consider a $d$-dimensional parameter space, where each dimension is divided into $N$ equiprobable sub-intervals. Let $i = 1, 2, \dots, N$ denote the sample index and $j = 1, 2, \dots, d$ represent the dimension index. For each dimension $j$, a random permutation $\pi_j$ is generated, such that $\pi_j(i)$ specifies the sub-interval containing the $i$-th sample in the $j$-th dimension. The actual sample value in the $j$-th dimension is then determined according to the following equation:

$$x_{ij} = \frac{\pi_j(i) - u_{ij}}{N} \tag{29}$$

In this context, $u_{ij}$ represents a random number drawn from the uniform distribution within dynamically determined spatial boundaries. This approach ensures that each sub-interval contains at least one sample while maintaining randomness within individual intervals. The spatial parameter predictor processes high-dimensional image features to infer dynamic spatial boundaries for the relevant physical domains along the $x$, $y$, and $z$ axes, represented as $[x_{min}, x_{max}]$, $[y_{min}, y_{max}]$, $[z_{min}, z_{max}]$. Subsequently, the LHS module utilizes these inferred three-dimensional spatial boundaries to generate a set of $N_S$ pseudo-random physical coordinates $(x_j, y_j, z_j)$. Compared with simple random sampling or uniform grid-based approaches, LHS achieves more uniform and comprehensive coverage of the three-dimensional physical space, thereby enhancing the robustness and accuracy of PINN physical loss calculations. Additionally, heat source parameters are predicted by dedicated heat source predictors (*Goldak_heat_source_pred* and *Conical_heat_source_pred*). The PINN then performs computations at each generated sampling point $(x_j, y_j, z_j)$, effectively integrating image-derived features into the physical consistency constraints of the model.

The pre-training stage of the PINN is illustrated in **Alg. 1**. A three-layer MLP is utilized to estimate the heat source parameters required for the loss function computation. To ensure a continuously differentiable neural representation, the Tanh activation function is adopted across intermediate layers, while the SoftPlus activation function is used in the output layer to enforce positivity of the predicted values.

**Alg. 1** Algorithm of pre-training stage.

**Input:** $D_{train}, D_{test}, D_{mem} \in D_U$, set of random transformations $F$, rotation degrees $R$, gaussian blurs $\Sigma$, crop images $C$, contrastive network $E_\theta, M_\theta$, PINN network $P_\theta$, learning rate $\alpha$

01: randomly initialize $\theta$
02: **for** $i$ in **epochs**
03: $b \leftarrow RandomSample(D_{train})$ //Select mini-batch samples
04: **for** $(x, t) \leftarrow b$ **do**
05: draw random rotation degrees $r_1, r_2 \sim R$
06: $x_1 \leftarrow r_1(x),\ x_2 \leftarrow r_2(x)$
07: draw random gaussian blurs $\sigma_1, \sigma_2 \sim \Sigma$
08: $x_1 \leftarrow \sigma_1(x_1),\ x_2 \leftarrow \sigma_2(x_2)$
09: draw random crop images $c_1, c_2 \sim C$
10: $x_1 \leftarrow c_1(x_1),\ x_2 \leftarrow c_2(x_2)$
11: draw random transformations $f_1, f_2 \sim F$
12: $x_1 \leftarrow f_1(x_1),\ x_2 \leftarrow f_2(x_1)$
13: $k_1 \leftarrow E_\theta(x_1), k_2 \leftarrow E_\theta(x_2), q_1 \leftarrow M_\theta(k_1), q_2 \leftarrow M_\theta(k_2)$
14: $L_{con} \leftarrow 1/2\ \|k_1, stopgrad(q_2)\|^2 + 1/2\ \|k_2, stopgrad(q_1)\|^2$
15: $(L_{Rot}, L_{Blur}, L_{crop}) \leftarrow 1/2\ \|x_1, r_1\|^2 + 1/2\ \|x_2, r_2\|^2$
16: $p_1 \leftarrow P(r_1), p_2 \leftarrow P(r_2)$
17: $spatial_params \leftarrow coord_pred(E_\theta(x), t)$
18: $sampled_coords \leftarrow LHS_module(spatial_params)$ // Generate $N_S$ sampling points in the physical space
19: $goldak_params \leftarrow Goldak_heat_source_pred(E_\theta(x), t)$
20: $conical_params \leftarrow Conical_heat_source_pred(E_\theta(x), t)$
21: $T \leftarrow temperature_pred(goldak_params, conical_params, spatial_params, t)$
22: $L_{PINN} \leftarrow 0$
23: **for** each sampled point $(x_j, y_j, z_j)$ in $sampled_coords$ **do:**
24: $\widehat{U}(t, x_j, y_j, z_j) \leftarrow P_\theta(goldak_params, conical_params, T_j, t, x_j, y_j, z_j)$
25: $L_{PDE_j} \leftarrow L_{PDE}(\widehat{U}, t, x_j, y_j, z_j)$
26: $L_{IBC_j} \leftarrow L_{IBC}(\widehat{U}, t, x_j, y_j, z_j)$
27: $L_{Con_j} \leftarrow L_{Con}(\widehat{U}, t, x_j, y_j, z_j)$
28: $L_{Gol_j} \leftarrow L_{Gol}(\widehat{U}, t, x_j, y_j, z_j)$
29: $L_{PINN} \leftarrow L_{PINN} + L_{PDE_j} + L_{IBC_j} + L_{Con_j} + L_{Gol_j}$
30: **end for**
31: $L_{PINN} \leftarrow L_{PINN}/N_S$

| | |
|---|---|
| 32: | $L \leftarrow L_{con} + \lambda_1 \cdot (L_{Rot} + L_{Blur} + L_{crop}) + \lambda_2 \cdot L_{PINN}$ |
| 33: | **end for** |
| 34: | $\theta \leftarrow \theta - \alpha \nabla_\theta L$ |
| 35: | $Acc \leftarrow KNN(E_\theta, M_\theta, P_\theta, D_{test}, D_{mem})$ |
| 36: | **end for** |
| 37: | **return** $E_\theta$ |

### 3.2 Fine-tuning stage

By leveraging the deep neural network architecture presented in Section 3.1, the proposed algorithm effectively extracts high-dimensional semantic features from laser molten pool images. Conventional approaches typically adopt parameter transfer learning strategies, fine-tuning pre-trained models on specific welding process datasets for weld penetration state classification. However, under limited training sample conditions, traditional fine-tuning methods are prone to overfitting, resulting in feature space distribution shifts and significant degradation in model generalization performance. To mitigate this challenge, this study introduces a novel algorithm based on the prototype network framework. The approach preserves the feature extraction capabilities of the pre-trained network while constructing class prototype vectors and establishing a metric space mapping mechanism, thereby enabling compact feature representation and maintaining generalization of classification boundaries in few-shot training scenarios.

Specifically, prototype networks learn a metric space $\Omega$, in which each class occupies a distinct region, with class centers maximally separated. For a new input sample $x$, the Euclidean distance between $x$ and each class center are calculated, and the class corresponding to the minimum distance is assigned as the predicted outcome. This underlying principle is illustrated in **Fig. 9**.

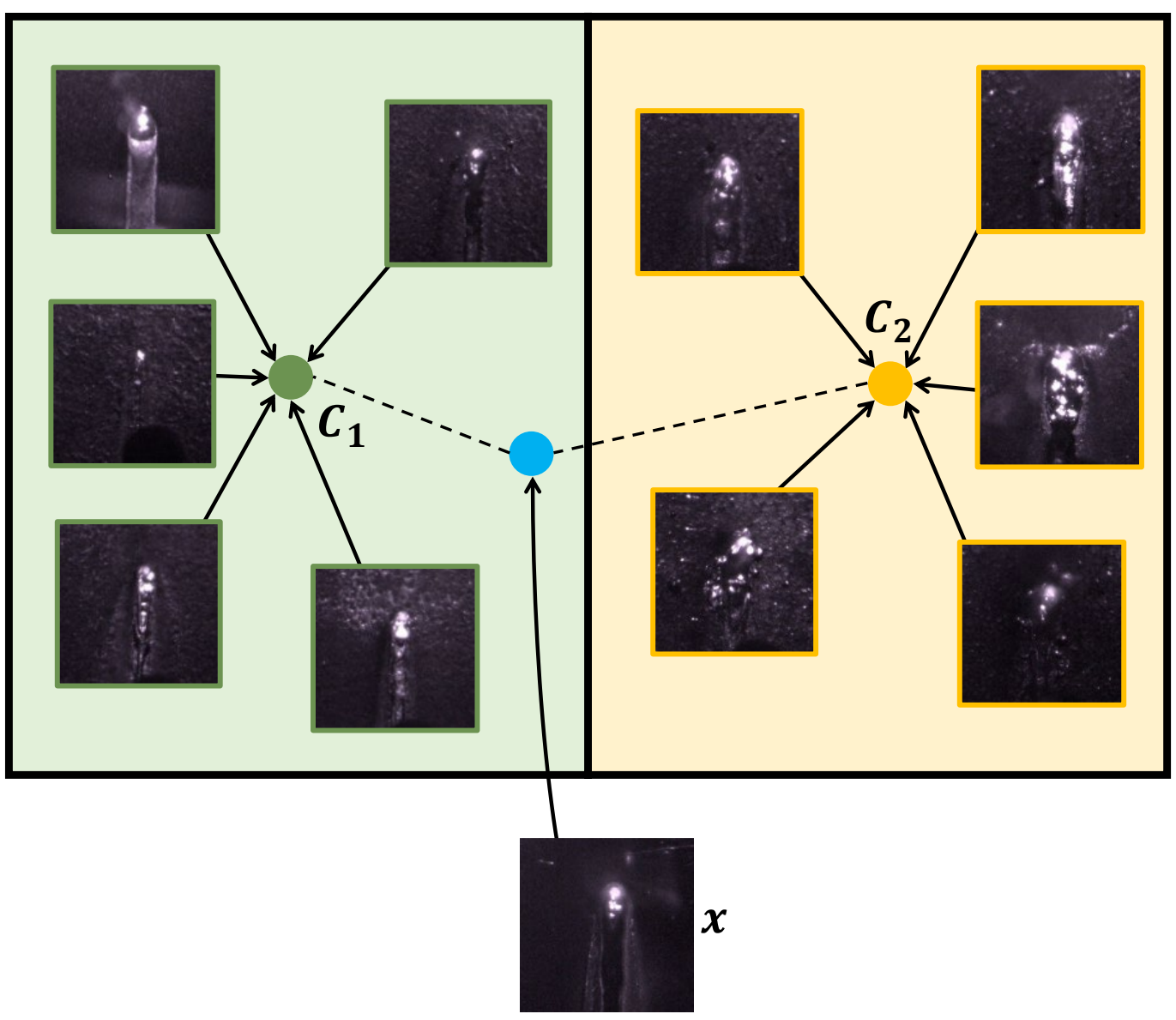

**Fig. 9** Prototypical network in a few-shot learning.

After completing the aforementioned two training stages, the proposed model demonstrates the ability to adapt to few-shot learning tasks. The training process during the fine-tuning stage is outlined in **Alg. 2**. As illustrated in **Alg. 2**, $N_s$ samples are first randomly selected from $D_L$ to form the support set $S_k$ and $N_Q$ samples are selected as the query set $Q_k$ (line 6).

For samples belonging to the same categories, it is assumed that the high-dimensional feature vectors extracted by the same neural network exhibit a certain level of similarity, whereas feature vectors from different categories may differ significantly. Therefore, the mean of the feature vectors corresponding to each category's samples is calculated and used as the support vector $c_k$, which can be mathematically expressed as:

$$c_k = (|S_k|)^{-1} \sum_{(x_i, y_i, t_i) \in S_k} E_\theta(x_i) \tag{30}$$

Consequently, to classify a molten pool image, its feature vector must first be calculated and compared to $c_k$ using the Euclidean distance (line 11). A smaller distance indicates a higher likelihood that the molten pool image belongs to that class. Therefore, the goal of few-shot learning is to ensure that support vectors effectively represent their respective classes by minimizing the distance between feature vectors sharing the same class label. The cross-entropy loss function is employed to update the network parameters (line 12).

In the fine-tuning stage, PINN also functions as an additional supervisory signal. In classification tasks, the model must ensure that its physical interpretation of the images is precise and consistent with physical laws. This approach enhances the model's generalization capability by enforcing that the learned classification features maintain physical consistency, thereby making the model's decisions more reliable and aligned with physical intuition when encountering previously unseen samples. The configuration of PINN is analogous to that outlined in **Alg. 1** and is not further elaborated here (line 20-29). The coefficient $\beta$ for the PINN network loss is set to 0.1 in our experiments.

**Alg. 2** Algorithm of fine-tuning stage.

**Input:** labeled training dataset $D_L = \{x_{i,L}, t_{i,L}, y_{i,L}\}_{i=1}^{N_L}$, Euclidean distance $d\,[\cdot,\cdot]$, learning rate $\alpha$

01: receive $E_\theta$ from pre-training stage
02: **for** $i$ in **epochs**
03: $N_{way} \leftarrow 2$ // Fix the number of categories *N* to 2
04: $L_{FSL} \leftarrow 0$
05: $L_{PINN} \leftarrow 0$
06: $(S_k, Q_k) \leftarrow RandomSample(D_L, N_{way}, K_{shot}, N_Q)$ // Select support sets and query sets
07: **for** $\boldsymbol{k}$ **in** $\{1, \ldots, N_{way}\}$ **do**
08: $c_k \leftarrow (|S_k|)^{-1} \sum_{(x_i, y_i, t_i) \in S_k} E_\theta(x_i)$
09: **end for**
10: **for** $(x_i, y_i, t_i)$ **in** $Q_k$ **do**
11: $d \leftarrow d[E_\theta(x_i), c_k] + log \sum_{k'} exp(-d[E_\theta(x_i), c'_k])$
12: $L_{FSL} \leftarrow L_{FSL} \cdot (|Q_k S_k|)^{-1} d$
13: **end for**

14: **for** $(X_i, Y_i, t_i)$ **in** $S_k$ **do**
15: $spatial_params \leftarrow coord_pred(E_\theta(X_i), t_i)$
16: $sampled_coords \leftarrow LHS_module(spatial_params)$
17: $goldak_params \leftarrow Goldak_heat_source_pred(E_\theta(X_i), t_i)$
18: $conical_params \leftarrow Conical_heat_source_pred(E_\theta(X_i), t_i)$
19: $T \leftarrow$
$temperature_pred(goldak_params, conical_params, spatial_params, t_i)$
20: $L_{PINN_i} \leftarrow 0$
21: **for** each sampled point $(x_j, y_j, z_j)$ in $sampled_coords$ **do**
22: $\widehat{U}(t_i, x_j, y_j, z_j) \leftarrow P_\theta(goldak_params, conical_params, T_j, t_i, x_j, y_j, z_j)$
23: $L_{PDE_j} \leftarrow L_{PDE}(\widehat{U}, t_i, x_j, y_j, z_j)$
24: $L_{IBC_j} \leftarrow L_{IBC}(\widehat{U}, t_i, x_j, y_j, z_j)$
25: $L_{Con_j} \leftarrow L_{Con}(\widehat{U}, t_i, x_j, y_j, z_j)$
26: $L_{Gol_j} \leftarrow L_{Gol}(\widehat{U}, t_i, x_j, y_j, z_j)$
27: $L_{PINN_i} \leftarrow L_{PINN_i} + L_{PDE_j} + L_{IBC_j} + L_{Con_j} + L_{Gol_j}$
28: **end for**
29: $L_{PINN_i} \leftarrow L_{PINN_i}/N_S$
30: $L_{PINN} \leftarrow L_{PINN} + L_{PINN_i}$
31: **end for**
32: $L_{PINN} \leftarrow L_{PINN_i}/(|S_k|)^{-1}$
33: $L \leftarrow \beta \cdot L_{PINN} + L_{FSL}$
34: $\theta \leftarrow \theta - \alpha \nabla_\theta L$
35: **end for**
36: **return** $E_\theta$

# 4. Experimental verification and analysis

## 4.1 Experimental results for pre-training stage

The experimental framework in this study is implemented using PyTorch, with the training environment configured on an Intel i7-13900KF CPU, an NVIDIA RTX 2080 Ti GPU, and 32 GB of RAM. The training configuration employs stochastic gradient descent (SGD)[21] as the optimizer, initialized with a learning rate of 0.001, and adopts the OneCycleLR scheduling strategy for dynamic learning rate adjustment. A batch size of 256 is used over 800 training epochs, and automatic mixed precision (AMP) is enabled to enhance computational efficiency. To evaluate model performance on downstream classification tasks, the K-nearest neighbors algorithm ($K = 2$)[22] is applied to assess classification accuracy based on high-dimensional features extracted from the trained model and test data. Specifically, the model is first trained on a laser welding penetration dataset comprising two categories, subsequently, KNN calculates Top-1 accuracy to quantify the discriminative capability of the learned features. This approach not only measures the model's effectiveness but also verifies the quality of the high-dimensional feature representations in supporting classification tasks. Additionally, ablation experiments are conducted to investigate the contribution of individual components within the proposed SimPhysNet architecture. The results of these ablation experiments are summarized in **Tab. 4** and illustrated in **Fig. 10**. Training dynamics, including accuracy and loss trajectories across epochs, are depicted in **Fig. 11**.

**Tab. 4** The ablation study of SimPhysNet. Best results are displayed in bold and the runner-up results are underlined.

| No. | Method | Acc(%)↑ |
|---|---|---|
| 1 | Simsiam | 83.15 |
| 2 | +Rot | 83.77(+0.62) |
| 3 | +Blur | 84.59(+0.82) |
| 4 | +Crop | 85.27(+0.68) |
| 5 | +Conical heat source | 87.43(+2.16) |
| 6 | + Goldak heat source | 88.17(+0.74) |
| 7 | +Muti-scales | 90.15(+1.97) |
| 8 | + PDE | 90.84(+0.69) |
| 9 | + IBC | 91.50(+0.66) |
| 10 | +Variable $\alpha$ | 92.04(+0.54) |
| 11 | No Latin | 90.88(-1.16) |
| 12 | points → 200 | 92.84(+1.96) |
| 13 | points → 500 | 93.68(+0.84) |
| 14 | points → 1000 | 94.55(+0.88) |
| 15 | Tanh → SoftPlus | <u>94.90(+0.34)</u> |
| 16 | Enhanced RMSE (SimPhysNet)(ours) | **95.14(+0.24)** |

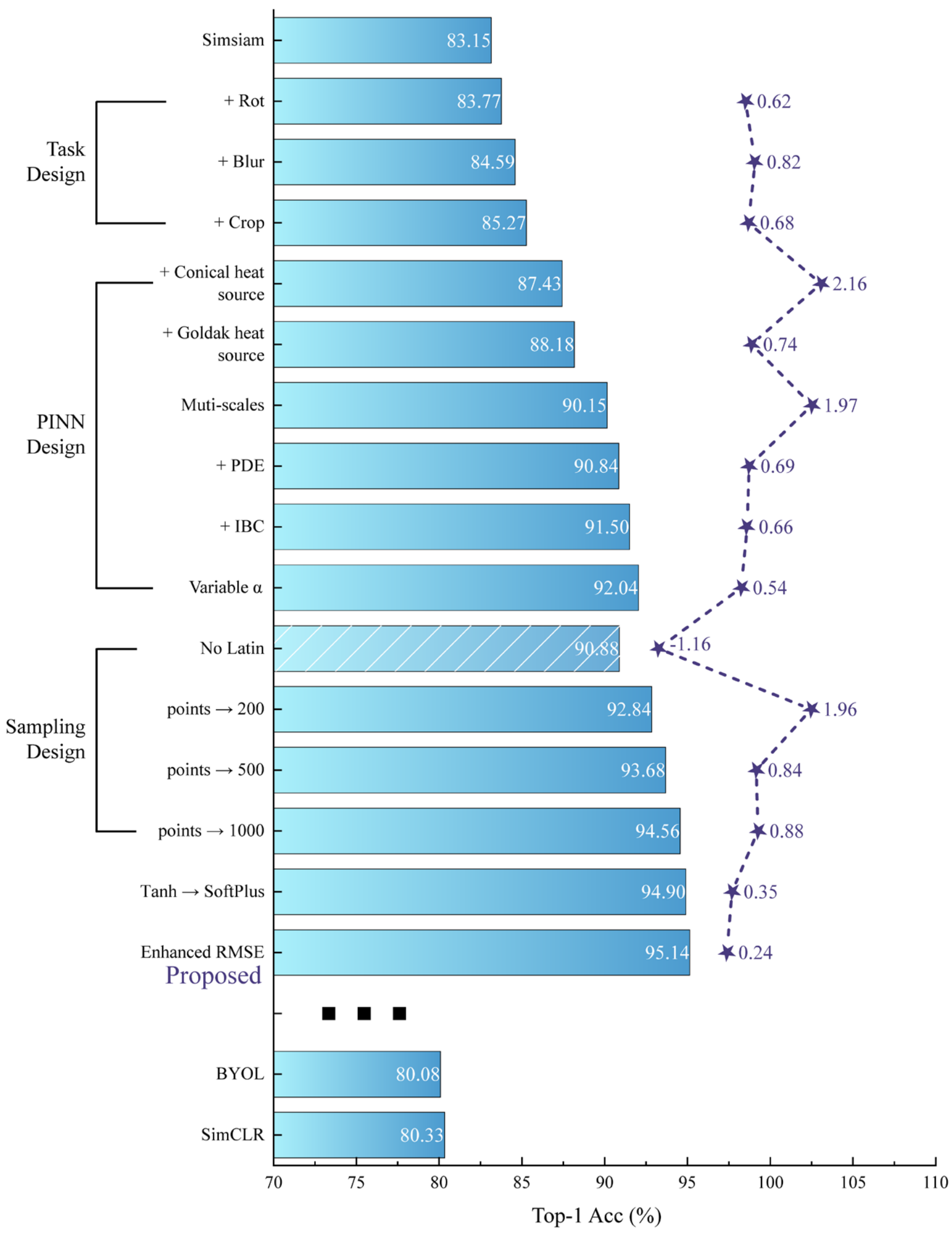

**Fig. 10** The ablation study of SimPhysNet. The bar chart represents model accuracy, the shaded bars represent unimplemented modifications, and the line graph represents the performance differences between ach modification and the preceding ones.

Experimental results indicate that the three image enhancement tasks proposed in this study contribute to varying degrees of performance improvement in the model. Notably, the Gaussian blur task achieves the most significant enhancement, which may be attributed to the prevalence of blurred images in the $D_U$ dataset, thereby underscoring the effectiveness of image augmentation strategies in improving model generalization. In the ablation experiments of the PINN framework, laser welding performance is found to be primarily influenced by the Conical heat source, yielding a 2.16% improvement in accuracy. The Goldak heat source also contributes to model performance, albeit to a lesser extent, with a marginal gain of 0.54%. Given that heat source dimensions may vary under different welding conditions and inspired by multi-scale architectures in deep learning, a multi-scale heat source approach—incorporating scales {0.25, 0.5, 1.0, 1.5, 2.0} is implemented to account for diverse heat source sizes across parameter settings, leading to a 1.77% performance gain. Furthermore, the inclusion of thermal conduction loss and boundary condition loss results in improvements of 0.69% and 0.66%, respectively. Considering that the material's thermal conductivity coefficient α is temperature-dependent in practical scenarios, a fully connected neural network layer is introduced to model the nonlinear relationship between thermal conductivity $\alpha$ and temperature $T$. Ablation experiments reveal that the proposed modification yields a 0.54% improvement in model performance. Notably, the exclusion of Latin hypercube sampling leads to a significant degradation in accuracy by -1.16%. The experimental results further indicate that increasing the number of sampling points enhances model accuracy; however, this also substantially increases model complexity. To achieve an optimal trade-off between predictive accuracy and computational cost, the number of sampling points is set to 1000 in this study. With regard to activation function optimization, replacing Tanh with SoftPlus ensures physically predictions—particularly non-negative values for heat source parameters resulting in an additional 0.35% performance gain. Furthermore, the improved RMSE algorithm contributes to more stable loss computation, resulting in a 0.24% improvement in overall model performance. Comparing the baseline Simsiam (No. 1 in **Tab. 4**, 83.15% accuracy) with the full SimPhysNet incorporating all PINN components (No. 14 in **Tab. 4**, 94.55% accuracy), the introduction of physical constraints directly yields a substantial improvement of 11.40% in classification accuracy. This significant gain underscores PINN's direct contribution to enhancing classification performance and implicitly improves model robustness by guiding feature learning towards physically consistent representations that are less prone to overfitting to spurious correlations, thereby better generalizing to real-world complexities.

Furthermore, we trained several contrastive learning models, and **Tab. 5** illustrates the accuracy of different contrastive learning algorithms on the $D_U$ dataset. The results indicate that the proposed pre-training method achieves higher accuracy, which can be attributed primarily to the constraints imposed by PINN on the model, as well as the enhancement of model generalization performance through three image augmentation tasks. These factors enable the pre-training stage to more effectively extract meaningful features from molten pool images. PINN can treat physical laws as a robust, unlabeled, and prior supervisory signal. In contrast to traditional unsupervised learning approaches, which mainly focus on uncovering the intrinsic structure of data, such as clustering or dimensionality reduction, without relying on external “labels” to guide the learning process, the core mechanism of PINN in this study involves loss function that integrates multiple residual terms derived from physical laws. The calculation of these residual terms does not depend on any labeled “ground truth” data; rather, it requires only that the outputs of SimPhysNet minimize these residuals. This self-supervised learning mechanism enables SimPhysNet to learn features that are both physically plausible and robust, even in the absence of explicit output labels.

**Tab. 5** Model accuracies of different networks. Best results are displayed in bold and the runner-up results are underlined.

| Method | Acc(%)↑ | P(%)↑ | R(%)↑ | F1(%)↑ |
|---|---|---|---|---|
| MoCo[23] | 76.46 | 71.15 | 80.07 | 75.37 |
| MoCov2[24] | 77.94 | 72.73 | 81.25 | 76.41 |
| BYOL[25] | 80.07 | 76.59 | 77.91 | 80.02 |
| SimCLR[26] | 80.33 | 77.18 | 79.61 | 79.09 |
| SimCLRv2[27] | 81.44 | 78.92 | 81.03 | 81.36 |
| Simsiam[13] | 83.15 | 80.97 | 85.38 | 83.12 |
| SimPhysNet (Ours) | **95.14** | **95.55** | **90.82** | **94.82** |

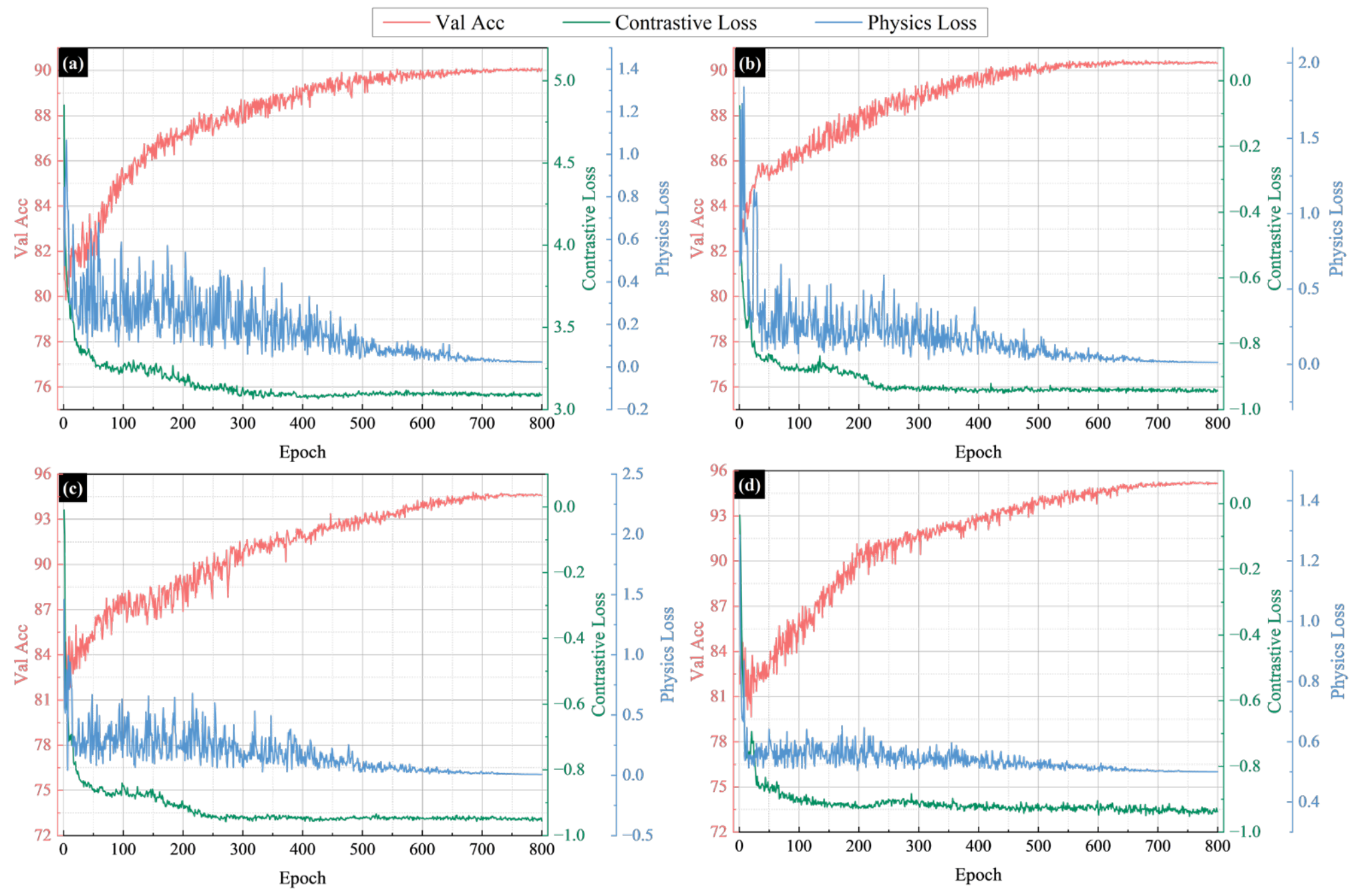

**Fig. 11** The losses and accuracies of SimPhysNet use KNN. (a) model No.8;(b) model No.9;(c) model No.14;(d) model No.16.

During the pre-training stage, Grad-CAM[28] was employed to visualize the trained model. It was observed that, as training progressed, the model gradually focused its attention on the molten pool and keyhole regions, demonstrating increasingly higher activation values in these areas relative to others. This suggests that the model effectively acquired the semantic features of molten pool images during pre-training. The corresponding attention maps are displayed in **Fig. 12**.

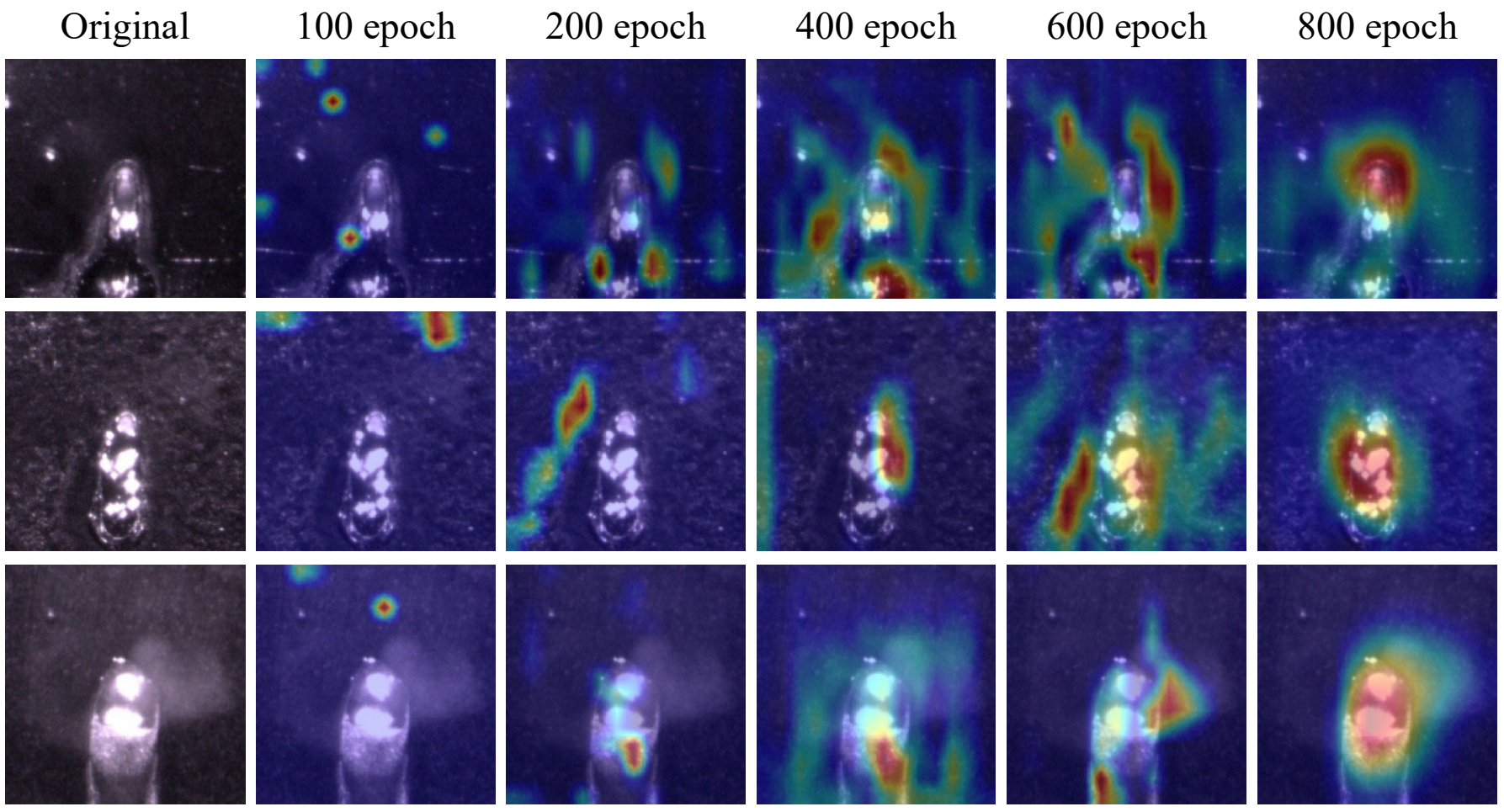

**Fig. 12** Activation map produced by Grad-CAM for SimPhysNet.

## 4.2 Experimental results for fine-tuning stage

Due to the limited number of categories in the laser welding penetration classification dataset, which consists of only two classes, namely penetrated and non-penetrated, this study evaluates the performance of SimPhysNet under 2-way 1-shot, 5-shot, 10-shot, and 20-shot settings. In each task, two test classes (2-way) are selected consistently, with each class comprising 1, 5, 10, and 20 support samples (K-shot), respectively, along with 1,000 query samples. To evaluated model performance, 1,000 unseen tasks were randomly sampled from the test classes. The performance of the proposed SimPhysNet was further compared with several commonly used few-shot learning (FSL) methods. Under a 95% confidence interval, **Tab. 6** reports the classification accuracies across the four aforementioned settings. As shown, SimPhysNet achieves the highest accuracy in all experimental configurations. Compared to the CSS method, SimPhysNet achieves an approximately 3% improvement. These results demonstrate that an effective pre-training strategy significantly contributes to model performance in few-shot classification tasks. Moreover, the observed performance gains validate the effectiveness of PINN in generating diverse feature representations within the embedding model, thereby enhancing generalization capabilities.

**Tab. 6** Performance comparison of the existing FSL methods. Best results are displayed in bold and the runner-up results are underlined.

| Method | 1-shot | 5-shot | 10-shot | 20-shot |
|---|---|---|---|---|
| MatchingNet[29] | 36.91 ± 0.63 | 62.75 ± 0.94 | 67.32 ± 0.85 | 72.96 ± 0.32 |
| ProtoNet[30] | 42.42 ± 0.74 | 70.20 ± 0.61 | 73.50 ± 0.70 | 76.89 ± 0.37 |
| RelationNet[31] | 53.44 ± 0.72 | 69.32 ± 0.57 | 72.00 ± 0.47 | 78.30 ± 0.30 |
| ProtoTransfer[32] | 38.74 ± 0.71 | 67.47 ± 0.81 | 72.46 ± 0.77 | 74.84 ± 0.31 |
| ProtoNet+jig[33] | 44.24 ± 0.77 | 70.04 ± 0.55 | 75.13 ± 0.63 | 77.37 ± 0.24 |
| ProtoNet+rot[33] | 57.78 ± 0.65 | 68.35 ± 0.69 | 73.81 ± 0.49 | 78.46 ± 0.15 |
| ProtoNet+rot+jig[33] | 59.33 ± 0.51 | 71.77 ± 0.67 | 76.43 ± 0.57 | 79.09 ± 0.21 |
| CSS[34] | 62.85 ± 0.84 | 72.08 ± 0.73 | 76.49 ± 0.39 | 80.59 ± 0.22 |
| SimPhysNet (Ours) | **65.30 ± 0.45** | **75.07 ± 0.36** | **80.10 ± 0.26** | **85.64 ± 0.18** |

Additionally, this study investigates the impact of support set size $N_s$ on classification accuracy. The model's performance is presented in **Fig. 13**, which illustrates the prediction accuracy as the support set size increases from 1 to 200. As shown in the figure, the classification accuracy improves progressively with increasing $N_s$, achieving 97.48% of the accuracy attained by the supervised learning model when $N_s$ reaches 200. When the support set is small, the model exhibits relatively lower classification accuracy. This phenomenon can be attributed to the fact that the prototype network calculates the distance between the features of the query image and the class prototypes derived from the support set. When the support set adequately covers the various states of the welding process, it more accurately represents the underlying distribution of sample images, thereby enhancing classification performance. Conversely, a limited support set leads to less reliable prototype estimation and consequently degrades accuracy. In summary, experimental results demonstrate that the proposed algorithm effectively learns discriminative features from molten pool images during the pre-training stage. Moreover, by leveraging the prototype network's few-shot fine-tuning strategy, the model achieves high classification accuracy even with a limited number of labeled samples. The training curves during the fine-tuning process are presented in **Fig. 14**

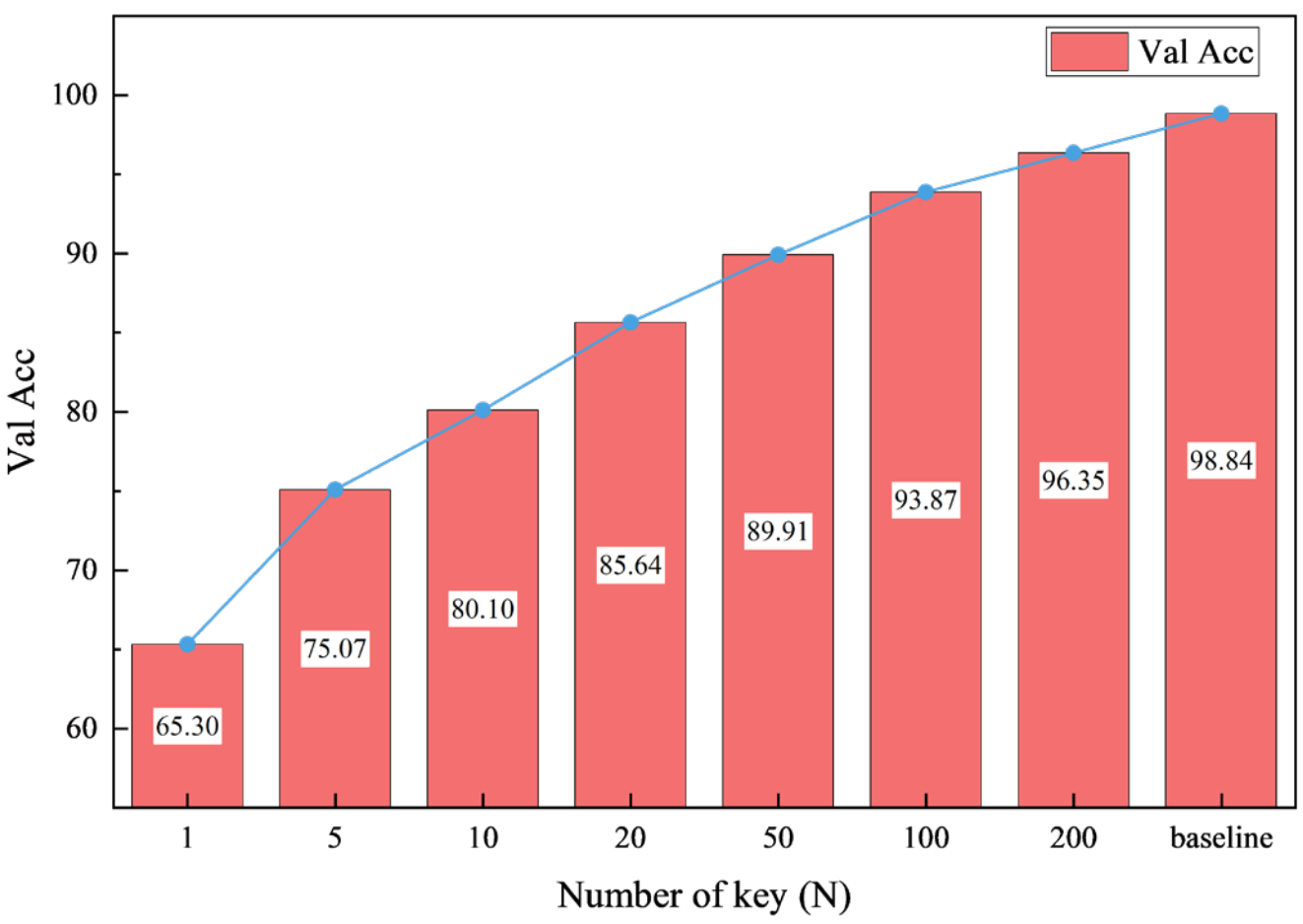

**Fig. 13** The results of different training-key N in SimPhysNet's fine-tuning stage.

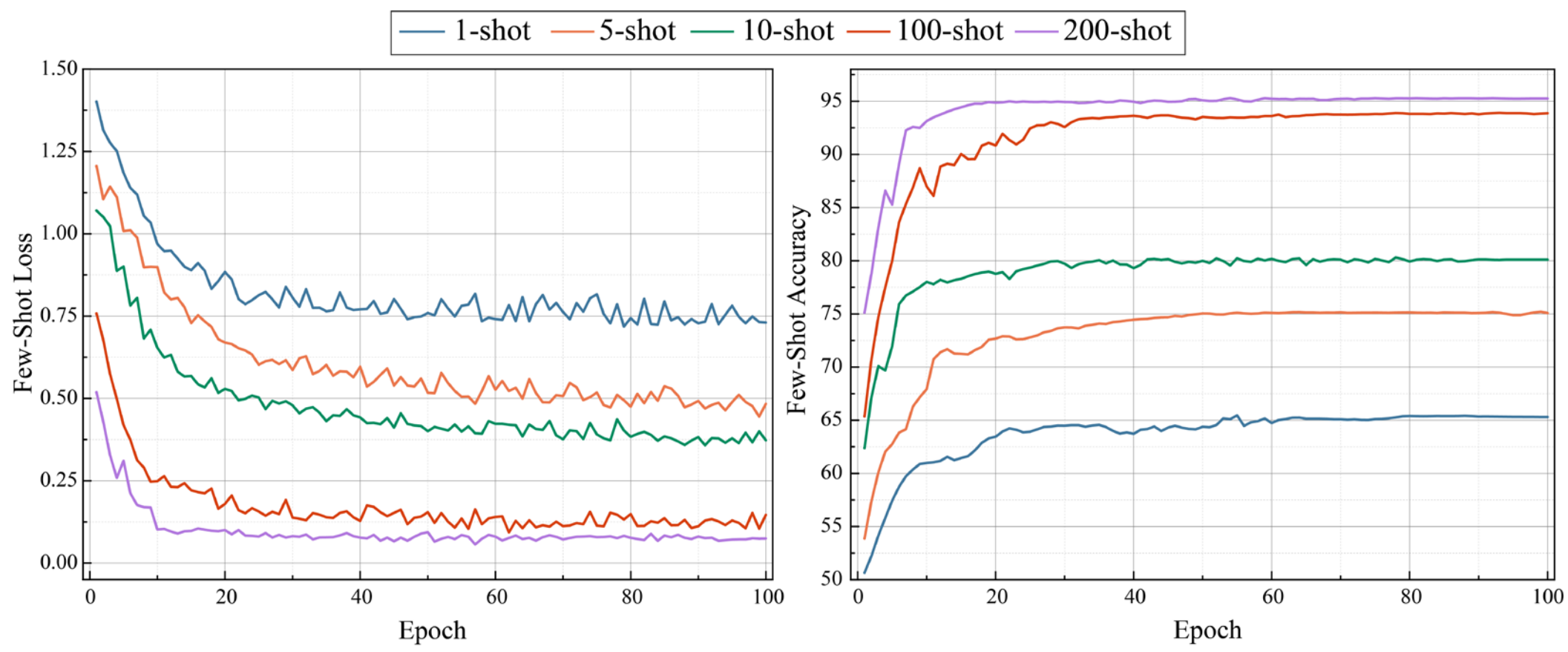

**Fig. 14** The losses and accuracies of SimPhysNet on 2-way.

### 4.3 Compare to supervised learning methods

The algorithm proposed in this paper aims to address the issue in supervised learning algorithms that rely on a large amount of high-quality labeled data. Therefore, it is necessary to compare the performance of both approaches. The supervised model is trained on the $D_L$ dataset, with data augmentation methods consistent with those used in SimPhysNet. The dataset is split into training and validation sets with a ratio of 7:3. The supervised learning model employed is ResNet-34, which achieves an accuracy of 98.84% after 100 training epochs. It is observed that the supervised model exhibits higher accuracy compared to SimPhysNet. Notably, the proposed contrastive learning pre-training combined with a few-shot fine-tuning strategy demonstrates excellent performance even when the number of labeled samples is significantly reduced: when the labeled data amounts to only 100 samples, the classification accuracy decreases by 4.97% to 93.87%; when the sample size increases to 200, the accuracy reaches 96.35% (a decrease of only 2.49%). These experimental results indicate that the feature representation space obtained through contrastive learning pre-training stage strong generalization capabilities. When combined with a few-shot learning strategy, it can approach the performance of supervised learning. This is particularly advantageous in industrial settings where labeled resources are limited, as it maintains high classification accuracy. Consequently, the method demonstrates substantial engineering application value in practical manufacturing scenarios.

The algorithm proposed in this paper addresses a critical limitation of supervised learning algorithms that rely on large volumes of high-quality labeled data. To evaluate its effectiveness, a comparative analysis of performance between the proposed approach and conventional supervised learning is conducted. The supervised model is trained on the $D_L$ dataset, with data augmentation techniques aligned with those utilized in SimPhysNet. The dataset is partitioned into training and validation subsets in a 7:3 ratio. A ResNet-34 architecture is adopted as the supervised learning model, achieving a classification accuracy of 98.84% after 100 epochs of training. The results demonstrate that the supervised model outperforms SimPhysNet in terms of accuracy. Notably, the proposed method—contrastive learning pre-training followed by few-shot fine-tuning—demonstrates robust performance even in scenarios with limited labeled data. Specifically, when only 100 labeled samples are available, the classification accuracy declines marginally to 93.87% (a reduction of 4.97%); with 200 labeled samples, accuracy reaches 96.35%, representing a decrease of merely 2.49%. These findings suggest that the feature representation space derived from contrastive learning pre-training possesses robust generalization capabilities. When integrated with a few-shot learning strategy, the method can closely approximate the performance of fully supervised learning. This characteristic is particularly beneficial in industrial environments where labeled data are scarce, enabling high classification accuracy with minimal annotation effort. Beyond accuracy, the practical utility of SimPhysNet hinges on its real-time performance. We measured the inference speed of the trained SimPhysNet model on an NVIDIA RTX 2080 Ti GPU The average inference time per molten pool image was found to be approximately 5.13ms, demonstrating its capability for rapid prediction. This high inference speed, combined with its robust performance on limited labeled data, confirms SimPhysNet's significant potential for real-time, online monitoring and control in laser welding processes.

UMAP [35] visualization was performed on both the supervised learning model and SimPhysNet, as depicted in **Fig. 15**. The figure illustrates that, as the number of pre-training epochs increases, SimPhysNet gradually separates the two categories of images into distinct feature clusters, achieving intra-class compactness within each cluster. The pre-training stage enables the model to effectively extract discriminative features from molten pool images. Consequently, during the subsequent fine-tuning stage with limited labeled samples, only minor adjustments to the network weights are required, allowing SimPhysNet to minimize intra-class distance while maximizing inter-class distance. These results indicate that SimPhysNet can accurately classify penetration states, thereby validating the effectiveness and reliability of the proposed approach.

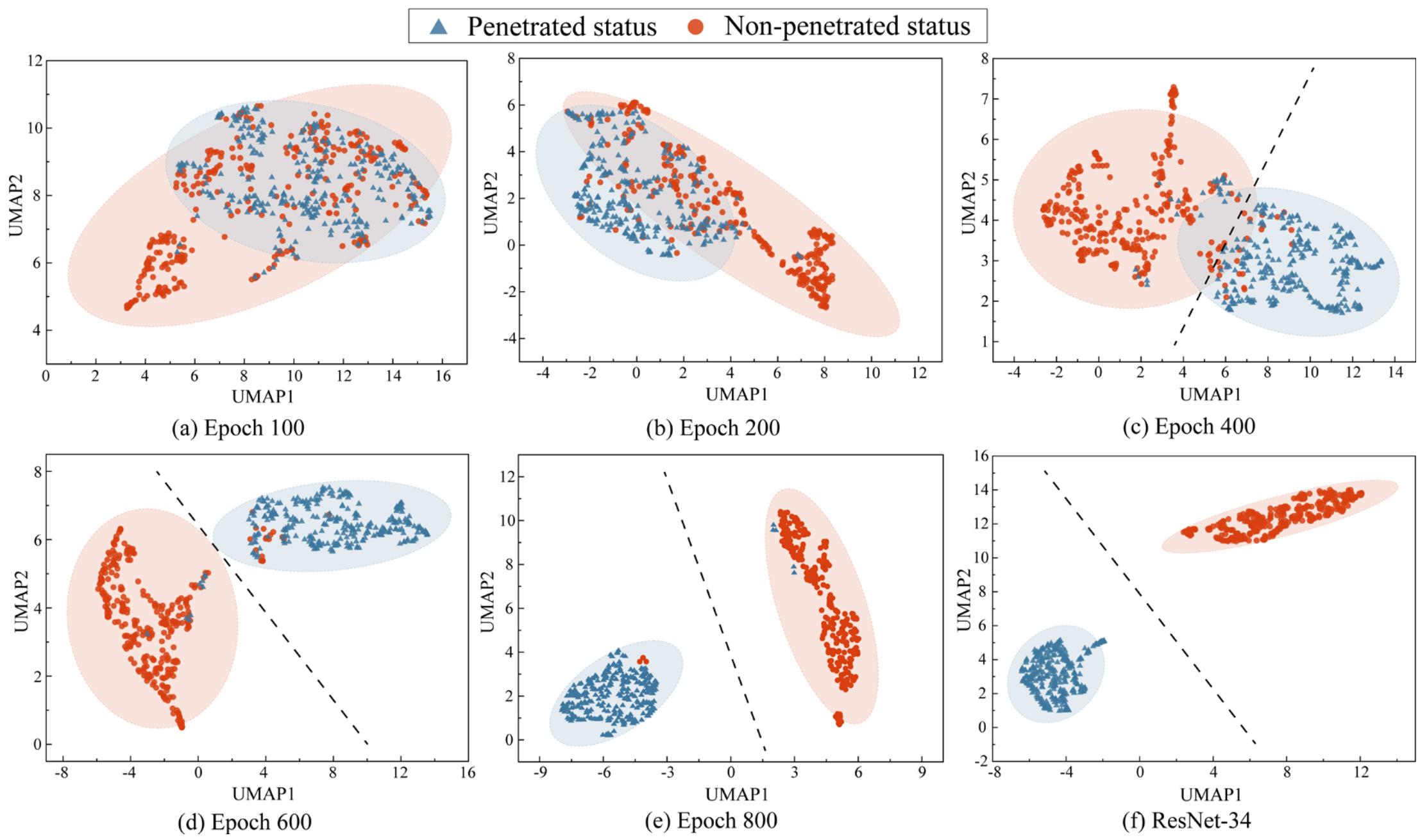

**Fig. 15** UMAP was utilized to visualize the output features of the model, with points of varying colors representing different classes. (a)-(e) depict the results of SimPhysNet under a 2-way 200-shot setting. (f) depicts the performance of the supervised learning (ResNet-34) on the training set. (For an explanation of the color references in the legend of this figure, please refer to the online version of this paper.)

**Fig. 16** presents the attention maps produced by the supervised learning algorithm and SimPhysNet. A comparative analysis of their attention distributions over the molten pool images reveals that both the keyhole and the molten pool regions play a critical role in the classification decisions made by the models. The high degree of similarity between the attention patterns generated by the two algorithms indicates that SimPhysNet effectively captures and learns salient visual features from the input images, demonstrating its capacity to emulate the behavior of fully supervised models.

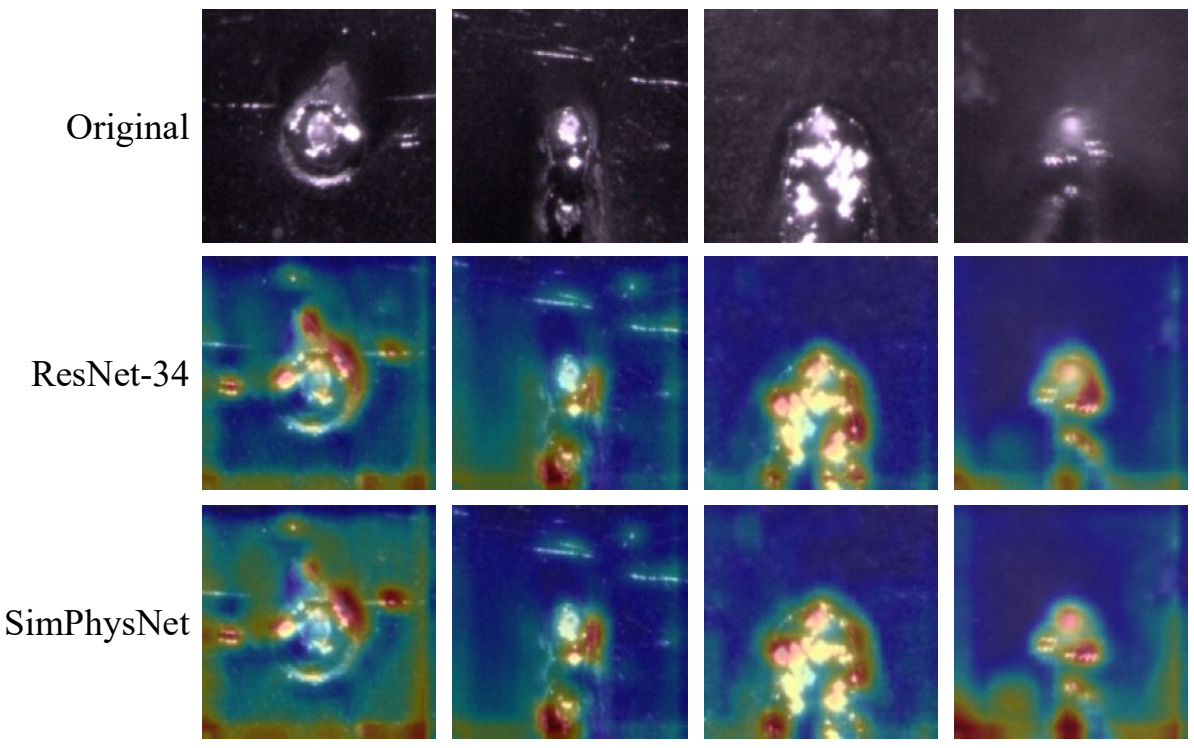

**Fig. 16** Activation map produced by Grad-CAM for supervised learning models, including ResNet-34 and SimPhysNet.

To more clearly demonstrate the performance across the three image enhancement tasks, we simultaneously visualized the results of all three enhancement methods. **Fig. 17** presents the attention maps produced by SimPhysNet under different rotation angles. It is evident that, regardless of the rotation angle, SimPhysNet consistently and accurately captures the molten pool, keyhole, and spatter regions. This suggests that SimPhysNet enhances the informational content of the images by integrating rotational transformations, thereby enabling more robust feature representations during training process.

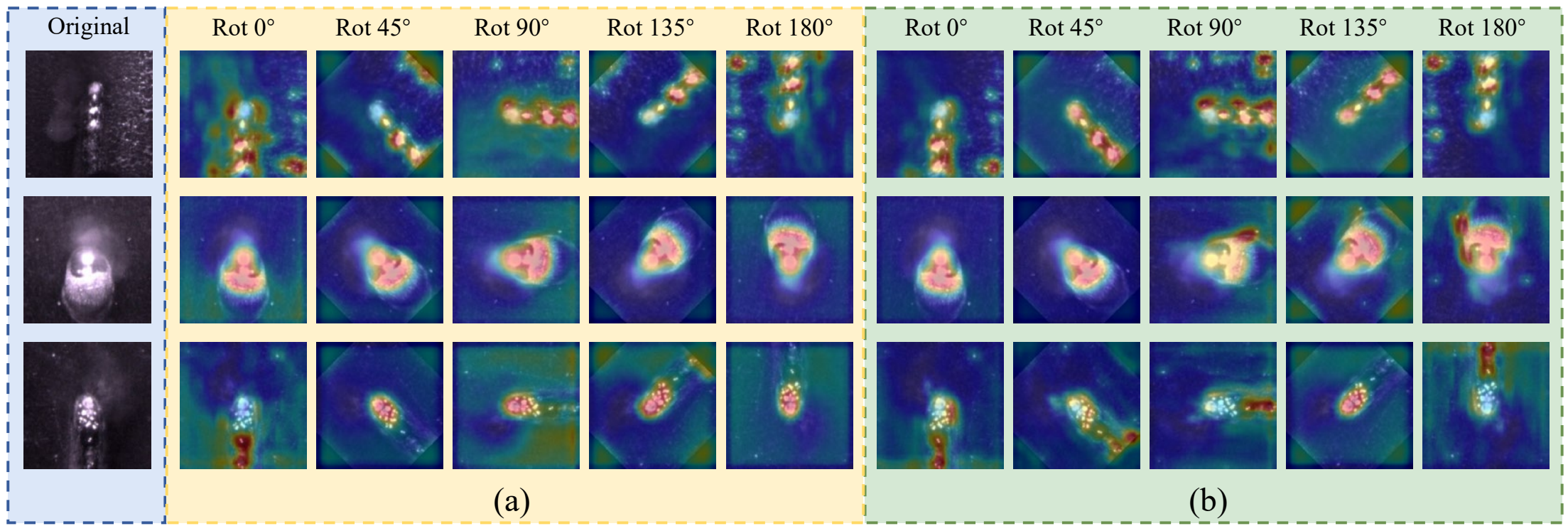

**Fig. 17** The results of the attention maps generated by Grad-CAM across various rotation angles. (a) Visualization results of SimPhysNet and (b) is generated by supervised learning (ResNet-34).

**Fig. 18** illustrates the attention maps generated by SimPhysNet under various gaussian blur parameters. The first row presents original images without gaussian blur, the second row shows results with different levels of artificially applied gaussian blur, and the fourth row displays images blurred during actual welding processes. Compared to the unblurred images, SimPhysNet exhibits minimal shift in attention localization across varying blur intensities. However, the spatial extent of attention gradually expands as the intensity of blurring increases. This expansion is attributed to the suppression of fine-grained details by gaussian blur, which leads the model to rely more heavily on global structural information. Notably, when applied to real-world welding images, SimPhysNet successfully captures regions containing critical features. These observations indicate that incorporating gaussian blur during training enhances the model's robustness and generalization capability in practical welding scenarios.

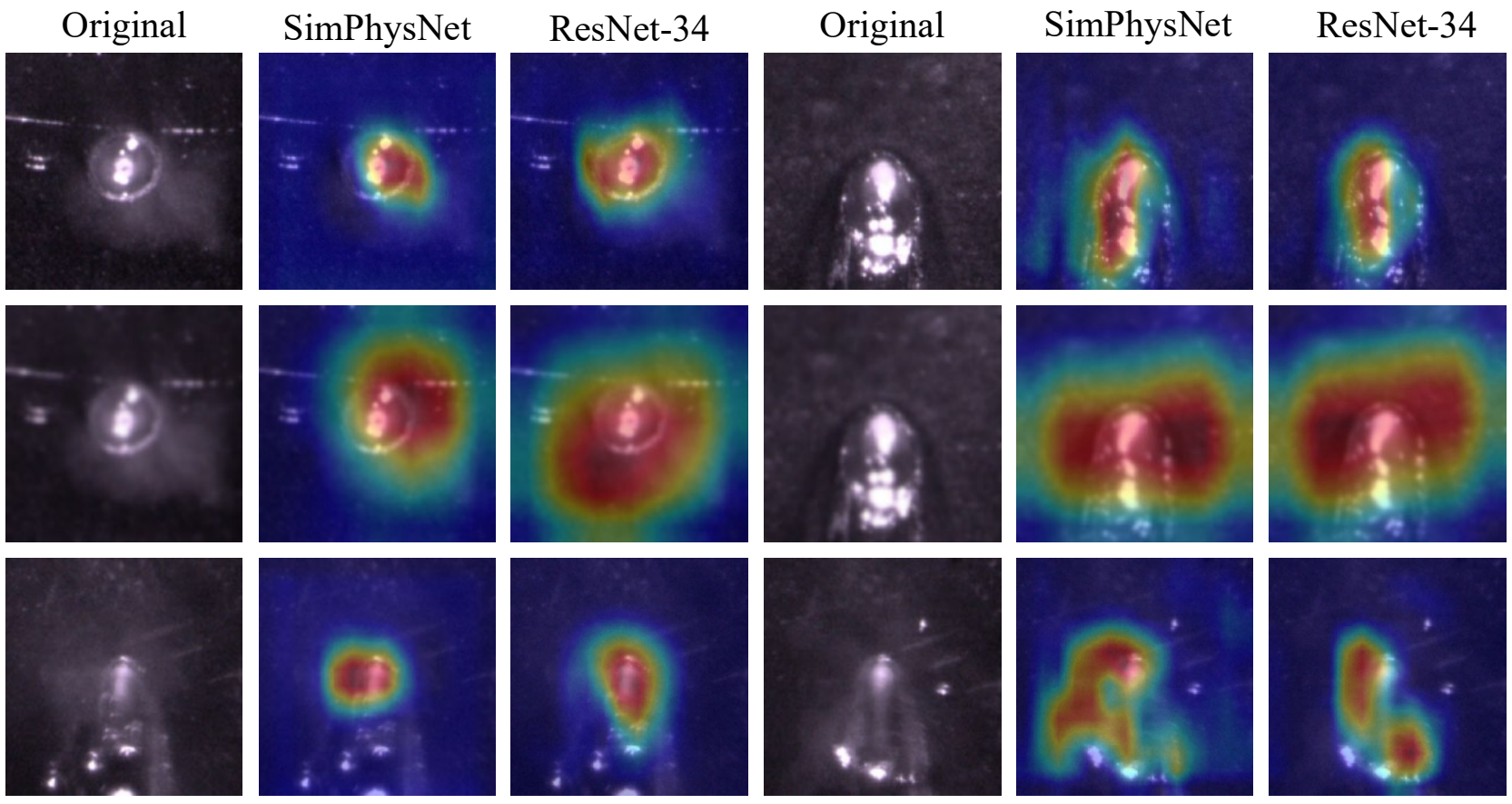

**Fig. 18** The results of attention maps generated by Grad-CAM under various Gaussian blur parameters using supervised learning (ResNet-34) and SimPhysNet.

**Fig. 19** presents the attention maps generated by our method under different cropping configurations. The figure comprises six images, each corresponding to a distinct cropping position. A comparative analysis of the attention maps reveals that image cropping encourages the model to extract discriminative features from fine-grained image details. Consequently, the model demonstrates heightened focuses on the molten pool boundaries and smaller-scale spatter regions.

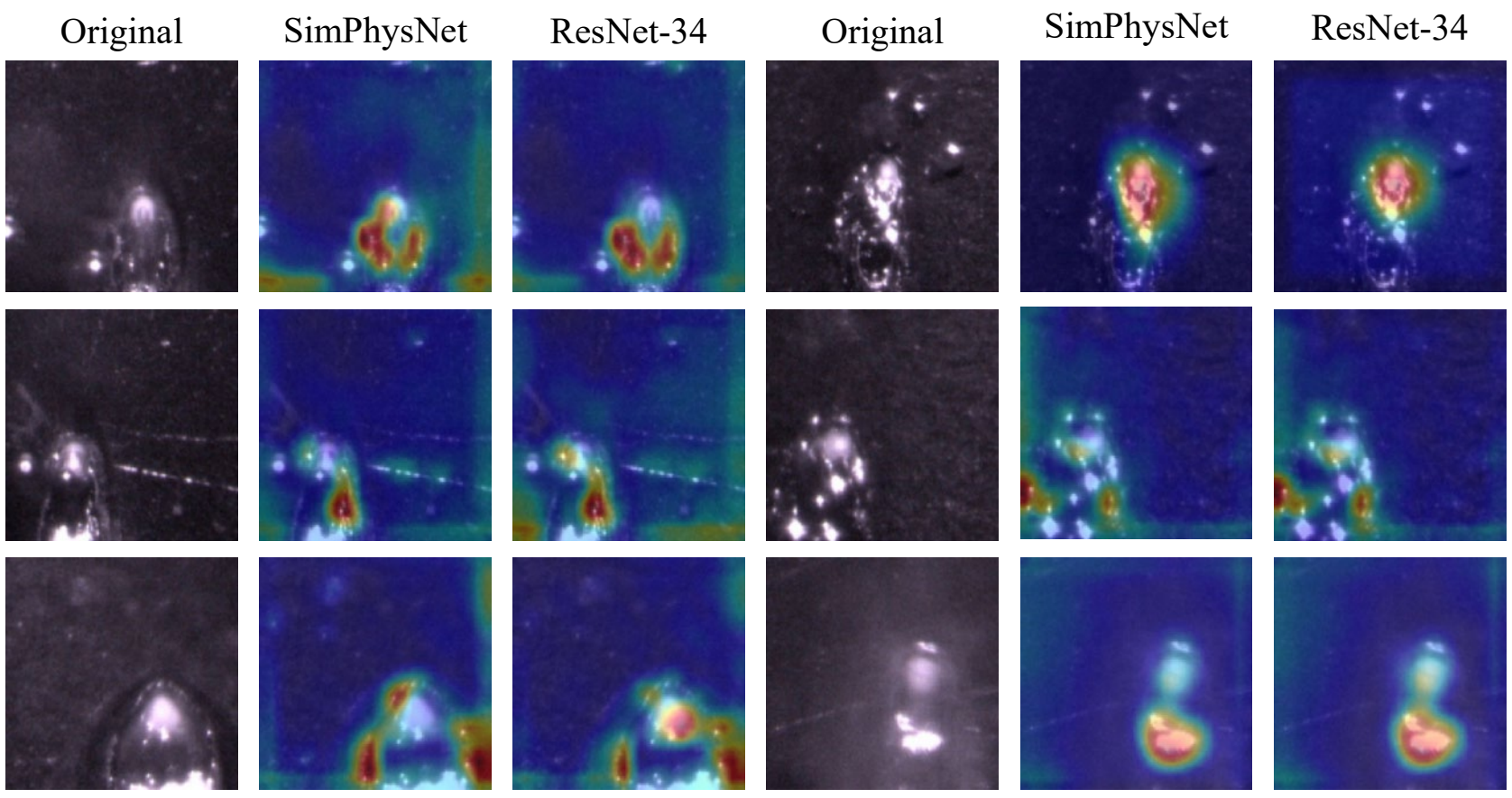

**Fig. 19** The results of attention maps generated by Grad-CAM at Various Cropping Positions using supervised learning (ResNet-34) and SimPhysNet.

Through the visualization analysis of the model attention (**Fig. 17**,**Fig. 18** and **Fig. 19**), we clearly demonstrate the advantages of SimPhysNet. The Grad-CAM results reveal its exceptional robustness achieved through image enhancement tasks. Whether the images are rotated, blurred, or partially occluded, the model accurately captures critical information. This series of visualization results compellingly confirms that the pre-trained framework employed by SimPhysNet effectively enhances the quality of feature extraction and the model's generalization performance, thereby ensuring its accuracy and reliability in practical welding applications.

Furthermore, to more effectively evaluate the performance of SimPhysNet, we conducted comparative experiments by training SVM[36], Random Forest[37], and XGBoost[38] models on the same $D_L$ dataset. The data augmentation and dataset partitioning strategies were kept consistent with those applied during supervi; sed learning. As shown in **Tab. 7**, the proposed training framework demonstrates superior feature representation capability. Moreover, compared to the baseline methods, the fine-tuning process achieves higher classification accuracy using only a limited number of labeled samples, which indicates that the pre-training stage effectively captures discriminative features. Experimental results indicate that, in comparison to conventional machine learning approaches, the proposed method exhibits enhanced generalization ability and adaptability across different settings.

**Tab. 7** Model accuracies of different networks. Best results are displayed in bold and the runner-up results are underlined.

| Method | Pre-training features | Acc(%)↑ | P(%)↑ | R(%)↑ | F1(%)↑ |
|---|---|---|---|---|---|
| SVM[36] | - | 88.13 | 83.66 | 87.20 | 87.96 |
| Random forest[37] | - | 85.18 | 85.53 | 85.13 | 82.59 |
| XGBoost[38] | - | <u>86.22</u> | <u>90.85</u> | <u>89.54</u> | <u>88.88</u> |
| ResNet-34[14] | - | **98.84** | **97.87** | **97.62** | **98.53** |
| SVM[36] | √ | 81.45 | 86.11 | 90.04 | 90.48 |
| Random forest[37] | √ | 89.08 | 88.83 | 87.39 | 86.71 |
| XGBoost[38] | √ | 90.32 | 93.10 | 92.74 | 92.45 |
| SimPhysNet (Our proposed) | √ | **95.14** | **95.55** | **90.82** | **94.82** |

In summary, SimPhysNet demonstrates higher accuracy than other algorithms reported in the literature, particularly when trained on smaller datasets. This approach effectively mitigates the high annotation costs typically associated with deep learning methods and exhibits superior generalization performance and adaptability in data-limited scenarios. By leveraging a large corpus of unlabeled data through a physics-informed self-supervised framework, SimPhysNet achieves a classification accuracy of 96.35% using only 5% of the total labeled dataset. This performance, representing a marginal decline of just 2.49% relative to a fully supervised model, demonstrates that the vast amounts of unlabeled process data typically generated in industrial environments can be effectively utilized to construct highly accurate predictive models.

## 5. Conclusion

Laser welding penetration classification typically relies on a large volume of labeled data. To address this limitation, this paper proposes a deep learning algorithm, termed SimPhysNet, which integrates two key components: a contrastive learning-based pre-training feature extractor and a few-shot classification-based feature classifier. The main contributions are summarized as follows:

1) By embedding physical priors governed by partial differential equations into the contrastive learning loss function, a physics-informed feature extraction framework is developed. This approach incorporates physical constraint terms to guides the learned feature space toward alignment with the underlying energy transfer dynamics of the actual welding process. Consequently, the model achieves enhanced prediction accuracy and improved interpretability, thereby increasing the reliability of its outputs.
2) Design of three image augmentation strategies for pre-training: Three supervised pretext tasks are introduced during the pre-training stage, which are rotation prediction, Gaussian blur level prediction, and random cropping localization. Rotation and blur prediction tasks encourage the network to capture global structural features, while random cropping localization promotes sensitivity to fine-grained details in weld pool images, such as melt pool boundaries. Experimental results demonstrate that these tasks effectively facilitate the contrastive learning process, enabling the network to extract discriminative features from molten pool imagery and improving performance in downstream classification tasks.
3) Building upon the concept of prototype networks, a laser welding penetration classifier was developed. This classifier utilizes the robust feature space established during pre-training to compute class-specific prototypes. By employing metric-based inference, it attains exceptional few-shot classification performance. This is demonstrated by the model's high accuracy across multiple scenarios, achieving 65.30% in a 2-way 1-shot setting. More notably, when trained with only 5% of the labeled dataset, the model maintains an accuracy of 96.35%, which is highly comparable to its fully supervised counterpart. his finding suggests a viable pathway to circumvent the data annotation bottleneck, substantially reducing the cost and effort required to deploy intelligent monitoring systems in practical manufacturing scenarios. The ability to approach supervised performance with minimal labeled data signifies the considerable engineering application value and scalability of the proposed methodology.

## Declarations

**Ethics approval** Not applicable.

**Consent to participate** The authors declare their consent to participate.

**Consent for publication** The authors declare their consent for publication.

**Conflict of interest** The authors declare no competing interests.

## Funding

The authors gratefully acknowledge the financial support from the National Key R&D Program of China (No. 2023YFB3407800), National Natural Science Foundation of China (No. U2141213).